\documentclass[10pt]{article}
\PassOptionsToPackage{table}{xcolor}
\PassOptionsToPackage{hyphens}{url}
\usepackage[accepted]{rlj}
\usepackage{amssymb}
\usepackage{mathtools}
\usepackage{graphicx}
\usepackage{booktabs}
\usepackage{microtype}
\usepackage{multirow}
\usepackage{algorithm}
\usepackage{algorithmic}
\usepackage{caption}
\usepackage{adjustbox}
\usepackage{longtable}
\usepackage{placeins}

\usepackage{amsfonts,bm}

\def\eqref#1{equation~\ref{#1}}

\def\1{\bm{1}}

\DeclareMathAlphabet{\mathsfit}{\encodingdefault}{\sfdefault}{m}{sl}
\SetMathAlphabet{\mathsfit}{bold}{\encodingdefault}{\sfdefault}{bx}{n}

\newcommand{\resulttableformat}{\small\renewcommand{\arraystretch}{1.08}\setlength{\tabcolsep}{2pt}}
\newcommand{\appendixtableformat}{\small\renewcommand{\arraystretch}{1.05}\setlength{\tabcolsep}{2pt}}
\newcommand{\ogtask}[2]{#1 / task#2}
\title{ReBRAC-v2: The Return of the King}
\setrunningtitle{ReBRAC-v2: The Return of the King}
\author{Denis Tarasov and Robert K.~Katzschmann}
\emails{denis.tarasov@srl.ethz.ch}
\affiliations{Soft Robotics Lab, D-MAVT\\ETH Zurich, Switzerland}

\begin{document}
\maketitle
\begin{abstract}
Recent offline reinforcement learning methods increasingly rely on expressive generative policies and specialized value guidance. We ask how far systematic modernization of a conventional behavior-regularized actor-critic can go while preserving algorithmic simplicity. \textbf{ReBRAC-v2} directly trains an exact-likelihood normalizing-flow actor with likelihood, MSE, and MAE behavior regularization, classification-based residual critics, staged optimization, and multi-sample inference. We develop one shared configuration through roughly 600 Bayesian proposals on six OGBench tasks, then adapt only two behavior-regularization coefficients using a 16-point grid for the main comparisons. Across ten common OGBench categories, ReBRAC-v2 averages 74.8 versus 52.3 for the next-best method and ranks first in eight categories. Without structural changes, it also obtains the strongest compared averages on D4RL AntMaze (90.2) and Adroit (33.6). Motivated by these results, we extend evaluation to all 235 state-based single-task problems across the 47 dataset variants in the original OGBench suite. To our knowledge, this is the first complete evaluation in this setting; no compared baseline reports results across the full suite. Per-task scores and configurations establish ReBRAC-v2 as a reference baseline for future work beyond the common task subset. The complete cloning objective retains the highest average after coefficient retuning of its alternatives, while further ablations characterize training, capacity, and inference choices. These results demonstrate the strength of systematic engineering within a minimalist offline RL foundation.

\unskip\footnote{Source code: \url{https://github.com/DT6A/ReBRAC-v2}}
\end{abstract}
\section{Introduction}
Offline reinforcement learning (RL) learns policies from a fixed dataset without further environment interaction \citep{levine2020offline}. This setting is attractive when online exploration is expensive or unsafe, but it makes policy improvement brittle: actions outside the dataset support can receive unreliable value estimates, and the learner cannot repair these errors by collecting new experiences. Offline RL methods must therefore improve over the behavior policy while controlling the distribution shift induced by that improvement.

Recent progress has increasingly relied on expressive generative policies and specialized mechanisms for guiding them with learned value functions. These methods can model multimodal behavior distributions but often introduce iterative action generation, auxiliary policies, distillation, or additional guidance machinery. At the same time, empirical performance depends heavily on less conspicuous choices in architecture, optimization, regularization, and evaluation. This creates an important methodological question: are increasingly elaborate offline RL algorithms \textit{necessary}, or can a conventional behavior-regularized actor-critic remain competitive when its design choices are systematically modernized?

ReBRAC \citep{tarasov2024revisiting} pursued the latter direction. Starting from TD3+BC \citep{fujimoto2021minimalist}, it showed that a carefully selected collection of practical modifications can substantially strengthen a behavior-regularized baseline without changing its basic algorithmic principles. We follow the same philosophy and ask how far a second generation of design choices can advance ReBRAC. Here, \emph{minimalism} refers to the algorithmic structure: ReBRAC-v2 remains a conventional actor-critic trained by Bellman updates, policy improvement, and behavior regularization. It does not imply the smallest networks, the least computation, or the absence of sophisticated function approximators.

\begin{figure}[t]
\centering
\includegraphics[width=0.51\linewidth]{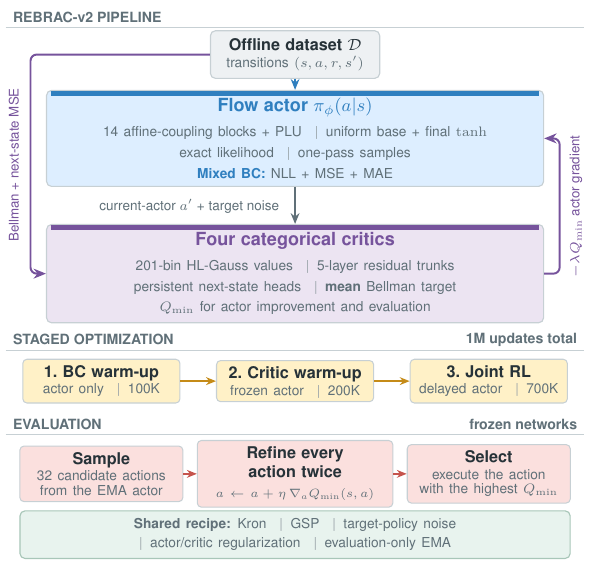}
\caption{ReBRAC-v2 pipeline. Mixed behavior cloning and $Q_{\min}$ policy gradients train the flow actor, while four categorical critics use a mean Bellman target. The lower strips summarize staged training and frozen-network inference.}
\label{fig:rebrac-v2-schema}

\end{figure}

The resulting \textbf{ReBRAC-v2} is not defined by one replacement but by the integration of complementary advances while preserving this structure. Building on NF-RLBC \citep{ghugare2025normalizing}, a major component is an \textit{exact-likelihood conditional normalizing-flow actor} trained directly by critic gradients and mixed behavior cloning. Unlike iterative diffusion or flow-matching policies, the flow itself is the end-to-end RL actor; no separate one-step policy is distilled from it. The mixed cloning objective combines flow likelihood with MSE and MAE attraction toward dataset actions. The critic, in turn, uses categorical value prediction, a deeper residual architecture, a compact ensemble, and auxiliary next-state prediction. Staged actor and critic warm-ups, selected optimization and regularization choices, and multi-sample critic-guided action improvement complete the recipe. Figure~\ref{fig:rebrac-v2-schema} summarizes how these components augment the ReBRAC foundation.

We also simplify the adaptation interface. ReBRAC regularizes both actor improvement and the critic target toward dataset actions. ReBRAC-v2 removes the critic-side penalty, whose effect was small in the original ablations and whose inclusion would introduce another environment-dependent coefficient. After developing one shared architectural, optimization, regularization, training, and inference recipe through Bayesian search on six challenging OGBench tasks \citep{park2024ogbench}, we freeze these choices. For the main comparisons, downstream adaptation changes only the likelihood and auxiliary cloning coefficients through an explicit $4\times4$ grid. On OGBench, the selected pair is transferred from one default task to the remaining tasks in the same environment. This fixed two-parameter interface changes far less than the released configurations of the recent GFP \citep{tiofack2025guided} and FAC \citep{chae2026flow} methods, as well as the concurrent DriftQL method \citep{houssaini2026drift}, which vary additional structural settings across categories or datasets.

\begin{figure}[t]
\centering
\begin{minipage}[b]{0.43\textwidth}
  \centering
  \includegraphics[width=\linewidth]{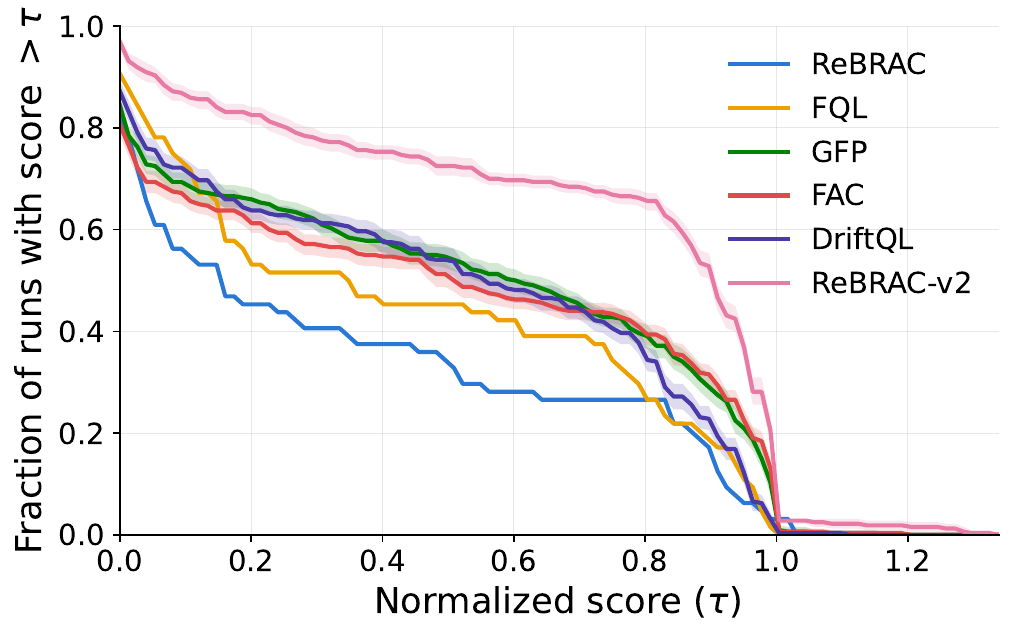}
  \par\vspace{2pt}\small (a) Performance profile.
\end{minipage}
\hfill
\begin{minipage}[b]{0.54\textwidth}
  \centering
  \includegraphics[width=\linewidth]{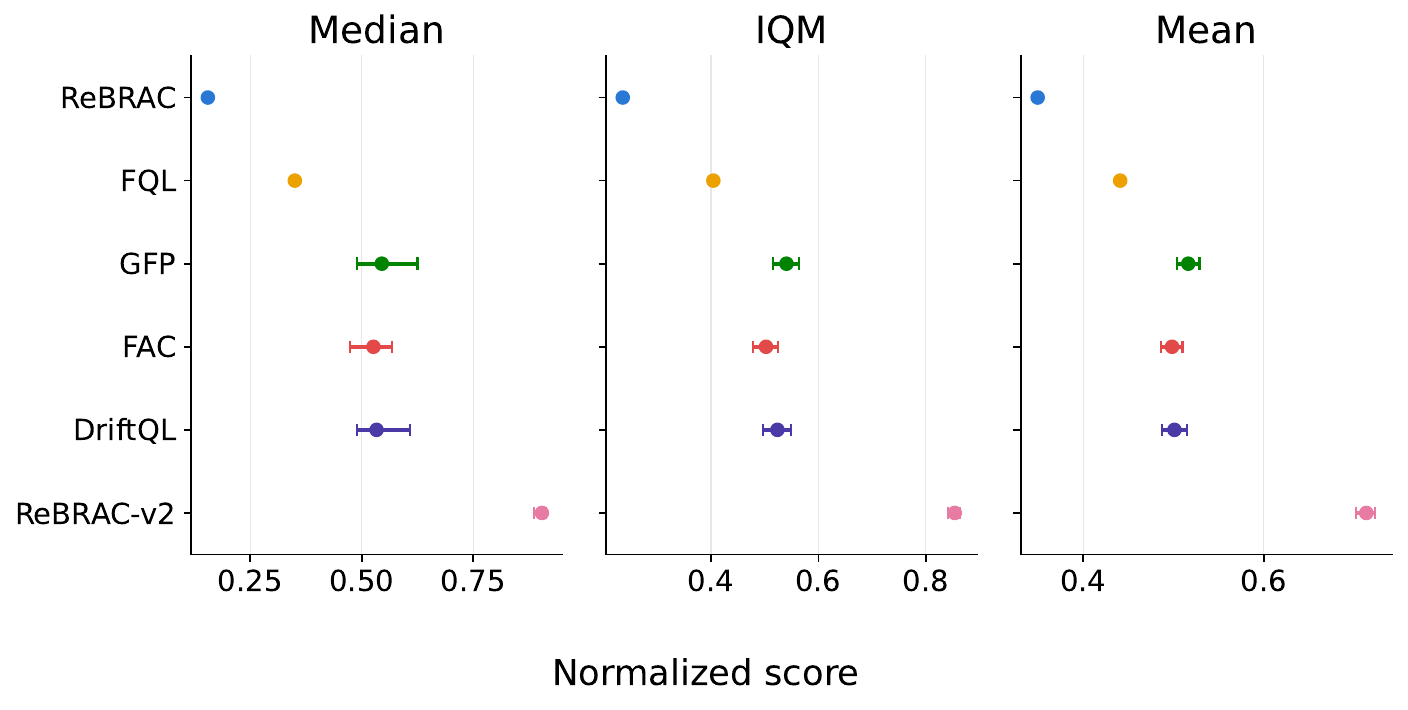}
  \par\vspace{2pt}\small (b) Aggregate normalized-score metrics.
\end{minipage}
\par\medskip
\begin{minipage}[b]{0.43\textwidth}
  \centering
  \includegraphics[width=\linewidth]{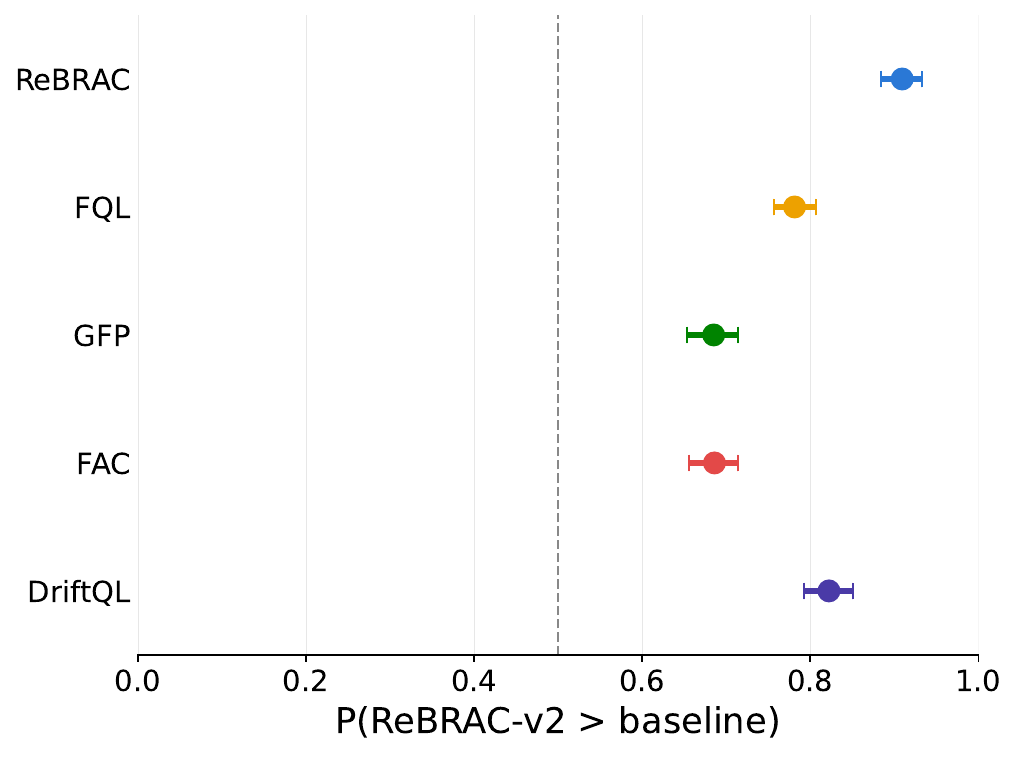}
  \par\vspace{2pt}\small (c) Probability of improvement.
\end{minipage}
\caption{Aggregate OGBench and D4RL performance. ReBRAC-v2 has the strongest performance profile, median, interquartile mean (IQM), and mean. Panel (c) estimates the probability that ReBRAC-v2 outperforms each baseline. Supplementary Section~\ref{sec:baseline-protocols} details uncertainty and score provenance.}
\label{fig:rliable-summary}

\end{figure}

As in ReBRAC, our contribution is systematic integration and controlled analysis, not the invention of individual ingredients. ReBRAC-v2 averages 74.8 across ten common OGBench categories versus 52.3 for the next-best method, leads in eight categories, and obtains the strongest compared D4RL AntMaze and Adroit averages of 90.2 and 33.6 \citep{fu2020d4rl}. Figure~\ref{fig:rliable-summary} shows broad improvements across both benchmarks. These results motivate evaluation beyond the common subset: we cover all 235 state-based single-task problems in the original OGBench suite. To our knowledge, this is the first complete evaluation in this setting. No compared baseline covers the entire suite; our per-task results and configurations provide a reference for future work on this broader set.

Our contributions are fourfold:
\begin{itemize}
    \item We systematically modernize ReBRAC while preserving its algorithmically minimalist, behavior-regularized actor-critic structure and evaluate the resulting components through fixed-recipe ablations and separate coefficient-retuned objective comparisons.
    \item Building on NF-RLBC's direct exact-likelihood flow actor, we combine critic-based policy improvement with mixed likelihood, MSE, and MAE behavior regularization without an iterative flow-matching policy or a distilled surrogate.
    \item We develop one shared recipe, use a 16-configuration two-coefficient adaptation grid for the main comparisons, and obtain the strongest compared aggregates on common OGBench tasks and D4RL AntMaze and Adroit.
    \item We evaluate the full original state-based single-task OGBench suite, providing scores and configurations across 47 dataset variants to support broader benchmarking beyond the common 50-task subset.
\end{itemize}

\section{Related Work}

\paragraph{Minimalist Offline RL.}
TD3+BC \citep{fujimoto2021minimalist} showed that a reconstruction penalty can turn TD3 \citep{fujimoto2018addressing} into a strong offline baseline, while ReBRAC \citep{tarasov2024revisiting} systematically improved that foundation through a set of implementation and design choices without changing its policy-improvement principle. IQL \citep{kostrikov2021offline} instead avoids evaluating unseen actions through expectile value learning and advantage-weighted imitation. ReBRAC-v2 continues the behavior-regularized line: it retains conventional Bellman learning and critic-guided policy improvement, removes the critic-side behavior penalty, and concentrates complexity in the actor representation, critic parameterization, optimization schedule, and evaluation procedure. Its contribution is therefore a carefully developed recipe rather than a new offline RL objective.

\paragraph{Generative Behavior Models.}
Diffusion and flow-matching policies can represent multimodal behavior distributions but ordinarily require iterative generation \citep{wang2022diffusion}. FQL \citep{fql_park2025} maintains an iterative flow-matching behavior policy and distills it into a separate one-step actor optimized by the critic. This separation makes fast execution possible while retaining an expressive behavior reference but introduces two policy parameterizations and a distillation relation between them. ReBRAC-v2 instead uses one discrete invertible normalizing flow as the RL actor itself. The same parameters receive exact-likelihood, pointwise cloning, and critic gradients, requiring neither numerical flow integration nor a distilled surrogate.

\paragraph{Recent Generative Offline RL.}
GFP \citep{tiofack2025guided} extends the two-policy FQL structure with mutual value-aware guidance between the flow behavior model and the distilled actor. FAC \citep{chae2026flow} uses a flow-matching proxy both to regularize a separate actor and to identify low-density actions for critic penalization. Concurrent DriftQL \citep{houssaini2026drift} takes a different one-pass route, learning a stochastic generator through kernel attraction, repulsion, and critic terms rather than an explicit-density flow. ReBRAC-v2 differs from all three by retaining a standard behavior-regularized actor-critic with one exact-likelihood flow actor and no generative proxy. At evaluation, it additionally samples, ranks, and refines actions with frozen critic gradients, building on test-time value optimization \citep{park2024value}.

\paragraph{Normalizing-Flow Actors.}
Normalizing flows provide reparameterized sampling, exact change-of-variables likelihoods, and one-pass generation. \citet{akimov2022let} pretrained a flow as a conservative action encoder and optimized a separate controller in its latent space. The closest actor formulation is NF-RLBC \citep{ghugare2025normalizing}, which already trains a conditional RealNVP/PLU actor directly through critic gradients and exact-likelihood behavior regularization in a minimalist RL+BC objective. We therefore do not claim the direct exact-likelihood flow actor or its likelihood-plus-$Q$ objective as standalone inventions. ReBRAC-v2 builds on this formulation by adding mixed MSE and MAE cloning, the modernized critic and training recipe, and test-time search. Supplementary Table~\ref{tab:ogbench-full} compares all methods on the 30 OGBench tasks reported for NF-RLBC. Related robotics results further motivate the policy parameterization \citep{tarasov2025nina,yang2026serfn}.

\paragraph{Value Learning and Optimization.}
Our critic follows evidence that categorical objectives can improve value estimation in deep RL and offline RL \citep{farebrother2024stop,tarasov2024value}. Its depth and residual structure are motivated by studies of scaling offline value functions and stabilizing deeper critics \citep{kumar2022offline,castanyer2025stable}. Actor regularization follows prior evidence that simple optimization regularizers materially affect offline RL \citep{tarasov2024role}; critic dropout and related regularization have likewise been studied as defenses against overestimation and instability \citep{hiraoka2021dropout}. We evaluate these components as interacting parts of the selected ReBRAC-v2 recipe, not as independent algorithmic contributions or universal rankings.

\paragraph{Ensembles, Training Schedules, and Averaging.}
Ensemble-based offline RL methods, such as EDAC, use critic diversity to control extrapolation error, but larger ensembles increase training costs \citep{an2021uncertainty}; ReBRAC-v2 therefore searches only two and four critics. Its warm-ups reflect the concern that a random critic should not immediately shape the policy; it is related to delayed policy updates and value pretraining \citep{fujimoto2018addressing,park2025pretraining}. We retain auxiliary transition prediction throughout training, avoiding another switch-time parameter. Evaluation-only actor EMA follows imitation-learning practice \citep{block2024butterfly}, but is not independently supported by our ablation. Thus, the search includes plausible stabilizers while our conclusions distinguish selection from controlled evidence.
\section{Preliminaries}

\paragraph{Offline RL and Behavior Regularization.}
We consider an MDP $\mathcal M=(\mathcal S,\mathcal A,P,r,\gamma)$ and a fixed dataset $\mathcal D=\{(s_i,a_i,r_i,s'_i)\}_{i=1}^{N}$ collected by unknown behavior policies \citep{levine2020offline}. A behavior-regularized actor-critic improves a policy using a learned action-value function while penalizing deviation from dataset actions \citep{wu2019behavior}. For $\tilde a\sim\pi_\phi(\cdot\mid s)$, a generic actor loss and Bellman target are
\begin{align}
\mathcal L_\pi(\phi)
&=\mathbb E_{(s,a)\sim\mathcal D}
\left[-Q_\theta(s,\tilde a)+\alpha_\pi F(\tilde a,a)\right],
\label{eq:generic-brac-actor}\\
y&=r+\gamma(1-d)
\left[Q_{\bar\theta}(s',a')-\alpha_Q F(a',\hat a')\right].
\label{eq:generic-brac-target}
\end{align}
Here $a'\sim\pi_\phi(\cdot\mid s')$ and $\hat a'$ are the corresponding dataset actions. TD3+BC and ReBRAC instantiate $F$ with simple action reconstruction penalties \citep{fujimoto2021minimalist,tarasov2024revisiting}. ReBRAC-v2 modifies both sides: it replaces the actor penalty with the mixed flow-based objective in Equation~\ref{eq:mixed-bc} and sets $\alpha_Q=0$, removing behavior regularization from the critic target.

\paragraph{Conditional Normalizing Flows.}
A normalizing-flow policy represents actions as an invertible state-conditioned transformation $a=f_\phi(z;s)$ of a simple base variable $z\sim p_0$ \citep{rezende2015variational,dinh2016density}. Its exact conditional likelihood follows from the change of variables:
\begin{equation}
\log\pi_\phi(a\mid s)
=\log p_0\!\left(f_\phi^{-1}(a;s)\right)
-\log\left|\det\frac{\partial f_\phi(z;s)}{\partial z}\right|.
\label{eq:flow-likelihood}
\end{equation}
Invertibility therefore provides both reparameterized policy samples and exact likelihoods for dataset actions. ReBRAC-v2 uses the same flow for critic-based policy improvement and likelihood-based behavior regularization.

\paragraph{Categorical Value Estimation.}
Following classification-based value estimation \citep{imani2018improving,farebrother2024stop,tarasov2024value}, each critic outputs logits over $K$ bins with centers $c_k$. The resulting scalar value is $Q_\theta(s,a)=\sum_{k=1}^{K}p_\theta(k\mid s,a)c_k$. A scalar Bellman target $y$ is projected to a soft categorical target $\Pi(y)$, and the critic minimizes
\begin{equation}
\mathcal L_{\mathrm{cls}}(\theta)
=-\mathbb E_{(s,a,y)\sim\mathcal D}
\left[\sum_{k=1}^{K}\Pi(y)_k\log p_\theta(k\mid s,a)\right].
\label{eq:categorical-critic}
\end{equation}
The Method specifies how ReBRAC-v2 constructs $y$, aggregates its critic ensemble, and augments this objective with transition prediction.
\section{Method}

\paragraph{Overview.}
ReBRAC-v2 retains the standard behavior-regularized actor-critic loop. A stochastic actor proposes an action, a critic ensemble supplies the policy-improvement gradient, and cloning terms constrain the actor toward the offline data. Figure~\ref{fig:rebrac-v2-schema} gives the high-level recipe; this section specifies how its components are parameterized and optimized. For $\tilde a\sim\pi_\phi(\cdot\mid s)$, the actor minimizes
\begin{equation}
\begin{aligned}
\mathcal L_{\mathrm{actor}}(\phi)
&=-\lambda\,\mathbb E_{s\sim\mathcal D}
\left[Q_{\min}(s,\tilde a)\right]\\
&\quad+\mathbb E_{(s,a)\sim\mathcal D}
\left[\mathcal L_{\mathrm{BC}}(\phi;s,a)\right],
\end{aligned}
\label{eq:full-actor-objective}
\end{equation}
where $Q_{\min}=\min_m Q_{\theta_m}$ and $\lambda=(\mathbb E|Q_{\min}|+10^{-6})^{-1}$ normalize the value scale. Each critic is trained with categorical Bellman prediction plus an auxiliary transition loss. Training first initializes the actor by behavior cloning, then learns the critic for the fixed actor, and finally alternates between critic and delayed actor updates. Supplementary Algorithm~\ref{alg:rebrac-v2} provides pseudocode.

\paragraph{End-to-End Normalizing-Flow Actor.}
We parameterize $\pi_\phi(a\mid s)$ as a conditional affine coupling flow \citep{dinh2016density,ghugare2025normalizing,tarasov2025nina,yang2026serfn}. The selected actor has 14 RealNVP-style coupling blocks, PLU dimension mixing adapted from invertible linear flow layers \citep{kingma2018glow}, a uniform base, and a final $\tanh$ transform. This provides exact likelihood, reparameterized one-pass sampling, and direct critic gradients; architecture details and search ranges appear in Supplementary Section~\ref{sec:method-details}.

For a dataset action $a$ and an independently sampled policy action $\tilde a$, the behavior-regularization term in Equation~\ref{eq:full-actor-objective} is
\begin{align}
\mathcal{L}_{\mathrm{BC}}(\phi;s,a)
={}&-\alpha_{\mathrm{NF}}
    \log \pi_{\phi}(a\mid s) \nonumber\\
&+\alpha_{\mathrm{aux}}
  \frac{1}{d_{\mathcal A}}
  \left(\lVert \tilde a-a\rVert_2^2
  +\lVert \tilde a-a\rVert_1\right).
\label{eq:mixed-bc}
\end{align}
where $d_{\mathcal A}$ is the action dimension; the implementation averages both auxiliary losses over action coordinates. The exact negative log-likelihood trains the full conditional density, while MSE and MAE directly attract policy samples toward dataset actions. Their gradients weight residuals differently: MSE emphasizes large deviations, whereas MAE retains a constant-magnitude correction for nonzero residuals. Their combination is empirical rather than derived as an optimal objective; the Ablation Studies section and Supplementary Table~\ref{tab:ablations-full} separately test the likelihood, MSE, and MAE components.

\paragraph{Categorical Residual Critic Ensemble.}
ReBRAC-v2 uses four critics. The current actor generates the next action $a'$, clipped target noise is added as in TD3, and the target critics are aggregated by their mean:
\begin{equation}
y=r+\gamma(1-d)\frac{1}{M}\sum_{m=1}^{M}
\bar Q_m(s',a'), \qquad M=4,
\label{eq:mean-ensemble-target}
\end{equation}
Each critic predicts 201 categorical value bins and minimizes cross-entropy toward the HL-Gauss projection $\Pi(y)$ from \citet{farebrother2024stop}, following its offline-RL use in \citet{tarasov2024value}. The search considers compact ensembles of two and four critics and minimum, mean, and maximum target aggregation; larger ensembles are excluded to control computation. The selected mean-bootstrap/minimum-improvement separation is empirical, and the joint two-critic/minimum-target ablation cannot attribute effects to ensemble size and aggregation independently.

Each critic has a five-layer residual trunk and separate two-layer value and next-state heads. GSP activation \citep{vitvitskyi2026mining} is selected over ReLU and SiLU. Motivated by next-state prediction for shared offline-RL value representations \citep{park2025pretraining}, the auxiliary head minimizes $\mathcal L_{\mathrm{dyn}}=\|g_\psi(h_\theta(s,a))-s'\|_2^2$ throughout training, encouraging dynamics-aware features without affecting rollouts or Bellman targets. Its small ablation effect makes it a secondary representation regularizer; Supplementary Section~\ref{sec:method-details} specifies the architecture and motivation.

\paragraph{Optimization, Regularization, and Training Schedule.}
The final configuration uses the Kronecker-factored PSGD optimizer Kron \citep{castanyer2025stable}; the search also includes AdamW \citep{loshchilov2017decoupled} and Adan \citep{xie2024adan}, motivated by evidence that optimization design materially affects deep RL \citep{castanyer2025stable,lan2023learning}. Competitive AdamW runs used different values for other hyperparameters, so this result does not establish a universal optimizer ranking. We search actor regularizers motivated by offline RL evidence \citep{tarasov2024role}, and empirically transfer analogous weight decay, gradient noise, and dropout choices to the critic; critic dropout also has direct precedent in offline RL \citep{hiraoka2021dropout}. Exact ranges appear in the supplementary material.

Training has three stages. First, 100,000 actor-only updates minimize Equation~\ref{eq:mixed-bc}, producing an initial behavior model before any critic gradient reaches the policy. Second, the actor is frozen for 200,000 critic-only updates, allowing value learning to begin under a fixed data-regularized policy. Third, the remaining budget jointly trains the critic and the full actor objective, with the actor updated every two critic steps. Both warm-up lengths are selected in Bayesian development. This schedule is motivated by SERNF \citep{yang2026serfn} and the general benefit of delaying policy updates until value estimates become more informative \citep{fujimoto2018addressing}. The two warm-up stages and subsequent joint stage together comprise the one-million-update training budget.

An actor EMA ($\tau_{\mathrm{EMA}}=0.005$) is used only for evaluation; critic targets use the current actor. Although common in imitation learning \citep{block2024butterfly}, EMA is retained only for protocol consistency because search and ablation evidence disagree.

\paragraph{Multi-Sample Test-Time Improvement.}
Following OPEX \citep{park2024value}, we use frozen value gradients to improve actions without updating network parameters. ReBRAC-v2 extends its single-action, single-step update by exploiting the flow actor's stochastic samples. At each state, we draw $K$ candidates $a_k^{(0)}\sim\pi_{\phi_{\mathrm{EMA}}}(\cdot\mid s)$ and refine each for $J$ steps:
\begin{equation}
a_k^{(j+1)}=\Pi_{\mathcal A}\!\left(
a_k^{(j)}+\eta
\frac{\nabla_a Q_{\min}(s,a)|_{a=a_k^{(j)}}}
{\lVert\nabla_a Q_{\min}(s,a)|_{a=a_k^{(j)}}\rVert_2+\epsilon}
\right),
\label{eq:test-time-refinement}
\end{equation}
where $Q_{\min}$ is the minimum prediction across the four critics, and $\Pi_{\mathcal A}$ clips the result to the valid action range. We then execute
\begin{equation}
a_{\mathrm{eval}}=\arg\max_{a\in\{a_k^{(J)}\}_{k=1}^{K}}Q_{\min}(s,a).
\label{eq:test-time-selection}
\end{equation}
The minimum across critics is used for both gradients and final selection, rejecting candidates that any critic evaluates as poor. No network parameters change during evaluation. Preliminary experiments fix $K=32$, $J=2$, and $\eta=0.01$ before Bayesian development; Figure~\ref{fig:inference-ablation} studies the sampling and refinement trade-off. 

\section{Experimental Setup}

\paragraph{Benchmarks.}
Our main comparison covers 50 state-based tasks shared by recent OGBench comparisons \citep{park2024ogbench}: five each from AntMaze Large/Giant, HumanoidMaze Medium/Large, AntSoccer Arena, Cube Single/Double, Scene, and Puzzle $3\times3$/$4\times4$. We additionally evaluate 185 tasks from 37 dataset variants, completing coverage of the original suite's 47 state-based variants and their five single-task problems each. This scope excludes visual observations and later large-scale dataset extensions. D4RL \citep{fu2020d4rl} contributes six AntMaze datasets and the human and cloned datasets for Pen, Door, Hammer, and Relocate. Scores are benchmark-defined and averaged within OGBench categories or D4RL suites. Detailed additional-task results and coefficients appear in Supplementary Section~\ref{sec:extended-ogbench}; no compared baseline covers this full set.

\paragraph{Shared Development and Adaptation.}
We separate development of a shared recipe from downstream behavior-regularization selection. W\&B Bayesian Sweeps\footnote{\url{https://docs.wandb.ai/models/sweeps/sweep-config-keys}} evaluates approximately 600 proposals on six challenging OGBench tasks, using one seed per task and maximizing their mean normalized score. For the main comparison, we then freeze all architectural, optimization, regularization, training, and inference choices and tune only $(\alpha_{\mathrm{NF}},\alpha_{\mathrm{aux}})$ over a 16-point grid using four seeds. On OGBench, the pair selected on an environment's default task is transferred to its other four tasks; on D4RL, a slightly shifted 16-point grid is selected per dataset. Exact tasks, grids, preliminary fixed choices, and search domains appear in the supplementary material.

\paragraph{Evaluation Protocol.}
We evaluate the final checkpoint after a fixed training budget. The main OGBench comparison uses five new seeds with 50 episodes per seed, while D4RL evaluation uses 10 new seeds with 100 episodes per AntMaze seed and 10 episodes per Adroit seed. ReBRAC-v2 seeds are disjoint from development and coefficient tuning.

\paragraph{Baseline Provenance.}
We rerun FAC \citep{chae2026flow}, GFP \citep{tiofack2025guided}, and DriftQL \citep{houssaini2026drift} with the authors' released implementations and hyperparameters under this protocol. In supplementary material we compare these reproductions with the corresponding source results.

\paragraph{Aggregate Analysis.}
Following RLiable \citep{agarwal2021deep}, Figure~\ref{fig:rliable-summary} reports the performance profile, median, interquartile mean, mean, and probability that ReBRAC-v2 improves over each baseline across the collected OGBench and D4RL tasks. We compute uncertainty intervals whenever per-seed results are available; the exact construction and its interpretation are detailed in the supplementary material.

\paragraph{Adaptation Scope and Selection Effects.}
ReBRAC-v2 exposes one fixed two-coefficient adaptation interface and does not change architecture, discount, aggregation, density estimation, or inference settings by downstream category or dataset. In contrast, released GFP, FAC, and DriftQL configurations vary additional structural choices; Supplementary Table~\ref{tab:released-tuning-envelopes} reconstructs these interfaces. The one-seed development scores are noisy and post-selection, but structural search covers only six of the ten OGBench categories. Evaluation additionally includes the unseen AntMaze Large, HumanoidMaze Medium, Cube Single, and Puzzle $3\times3$ categories, four held-out tasks within each developed category, and all D4RL datasets.

\begin{table}[!t]
\caption{Aggregate OGBench and D4RL results. OGBench rows average five tasks; bold and underline mark the best and second-best result. Supplementary Tables~\ref{tab:d4rl-full} and~\ref{tab:ogbench-full} give task-level scores, and Supplementary Section~\ref{sec:matched-baseline-reproductions} provides the source-to-reproduction comparison.}
\label{tab:main-results}
\centering
\resulttableformat

\begin{adjustbox}{max width=\linewidth}
\begin{tabular}{@{}lrrrrrrrrr@{}}

\addlinespace[2pt]
& \multicolumn{3}{c}{\textbf{Gaussian Policies}} & \multicolumn{4}{c}{\textbf{Flow Policies}} & \textbf{Drift} & \shortstack{\textbf{Normalizing}\\\textbf{Flow}}\\
\cmidrule(lr){2-4} \cmidrule(lr){5-8} \cmidrule(lr){9-9} \cmidrule(lr){10-10}
\multicolumn{1}{c}{\textbf{Task Category}} & \textbf{BC} & \textbf{IQL} & \textbf{ReBRAC} & \textbf{IFQL} & \textbf{FQL} & \textbf{FAC} & \textbf{GFP} & \textbf{DriftQL} & \textbf{ReBRAC-v2} \\
\midrule

    AntMaze Large & 10.6 & 53.4 & 80.8 & 28.0 & 78.6 & 57.2 & \underline{93.0} & 88.5 & \textbf{97.8} \\
    AntMaze Giant & 0.2 & 4.0 & 26.2 & 2.6 & 8.6 & 0.2 & 32.0 & \underline{61.2} & \textbf{75.0} \\
    HumanoidMaze Medium & 2.0 & 32.8 & 21.8 & 60.4 & 57.4 & 67.3 & \underline{70.4} & 61.4 & \textbf{87.0} \\
    HumanoidMaze Large & 0.4 & 2.4 & 2.6 & 11.0 & 4.2 & 4.0 & \underline{14.9} & 5.0 & \textbf{72.2} \\
    AntSoccer Arena & 1.0 & 8.4 & 0.0 & 33.2 & 60.2 & \underline{65.5} & 64.8 & 62.2 & \textbf{67.4} \\
    \midrule
    Cube Single & 5.4 & 83.0 & 90.6 & 79.2 & \underline{95.8} & 95.4 & \textbf{98.5} & 87.7 & 93.4 \\
    Cube Double & 1.6 & 6.4 & 12.2 & 14.0 & \underline{28.6} & 26.0 & \textbf{40.1} & 17.8 & 8.3 \\
    Scene & 4.6 & 27.6 & 40.6 & 30.4 & 55.8 & 60.9 & 53.7 & \underline{72.0} & \textbf{95.4} \\
    Puzzle 3x3 & 1.8 & 9.0 & 21.6 & 19.0 & 29.6 & \underline{98.8} & 23.3 & 40.2 & \textbf{99.6} \\
    Puzzle 4x4 & 0.2 & 7.4 & 14.0 & 25.2 & 17.2 & 26.5 & \underline{30.5} & 27.0 & \textbf{52.2} \\
    \midrule
    {\textbf{OGBench Average}} & 2.8 & 23.4 & 31.0 & 30.3 & 43.6 & 50.2 & 52.1 & \underline{52.3} & \textbf{74.8} \\
    \midrule
    \textbf{D4RL AntMaze Average} & 17.2 & 57.2 & 78.7 & 64.8 & 83.5 & \underline{85.3} & 83.3 & 77.3 & \textbf{90.2} \\
    \midrule
\textbf{D4RL Adroit Average} & 16.1 & 21.4 & \underline{26.6} & 20.6 & 17.6 & 21.9 & 21.4 & 12.5 & \textbf{33.6} \\

    \bottomrule
\end{tabular}
\end{adjustbox}

\label{tab:d4rl-results}
\label{tab:ogbench-results}
\end{table}

\section{Main Results}

\paragraph{OGBench.}
ReBRAC-v2 averages 74.8 across the ten common OGBench categories, 22.5 points above DriftQL's next-best aggregate of 52.3 (Table~\ref{tab:main-results}). It ranks first in eight categories; Figure~\ref{fig:rliable-summary} shows the strongest median, IQM, and mean and a performance profile dominating most thresholds. ReBRAC-v2 scores 75.0 on AntMaze Giant versus 61.2 for DriftQL, 72.2 on HumanoidMaze Large versus 14.9 for GFP, 95.4 on Scene versus 72.0 for DriftQL, and 52.2 on Puzzle $4\times4$ versus 30.5 for GFP. It reaches 99.6 on Puzzle $3\times3$ and remains competitive on Cube Single and AntSoccer Arena, demonstrating gains across multiple families rather than an isolated outlier.

\paragraph{Performance Beyond Selection Tasks.}
Table~\ref{tab:ogbench-selection-scope} separates the 50 common OGBench tasks by their role in ReBRAC-v2's documented selection procedure: six default tasks used for Bayesian development and coefficient selection, four additional default tasks used only for coefficient selection, and 40 non-default tasks excluded from both stages. On these 40 tasks, ReBRAC-v2 averages 74.1 versus 54.2 for the strongest baseline aggregate, DriftQL, a 20.0-point advantage before rounding. The advantage therefore persists when all selection tasks are excluded from the reported average.

Within this 40-task group, 16 tasks belong to the four categories absent from Bayesian development: AntMaze Large, HumanoidMaze Medium, Cube Single, and Puzzle $3\times3$. ReBRAC-v2 averages 93.5 on this subset versus 79.8 for FAC, the strongest baseline aggregate. These tasks were not used to select either the shared recipe or the category-specific coefficients; the coefficients were selected on their categories' default tasks. This supports transfer across task variants and to categories outside shared-recipe development, but is not a fully independent benchmark: tasks within a category share the offline dataset and environment, and the categories still receive coefficient tuning. These groups describe ReBRAC-v2's selection history, not a claim that the baselines had identical development splits.

\begin{table}[t]
\caption{OGBench results separated by ReBRAC-v2's documented model-selection scope. Scores equally average task-level means under the main final-checkpoint protocol. The first three rows partition the 50 main-comparison tasks; the final row is a 16-task subset of the 40-task row, not an additional disjoint group. Selection labels describe ReBRAC-v2, not the baselines' development histories.}
\label{tab:ogbench-selection-scope}
\centering
\small
\setlength{\tabcolsep}{3pt}
\renewcommand{\arraystretch}{1.10}
\begin{adjustbox}{max width=\linewidth}
\begin{tabular}{@{}p{0.44\linewidth}rrrrr@{}}
\toprule
\textbf{ReBRAC-v2 selection scope} & \textbf{Tasks} & \textbf{FAC} & \textbf{GFP} & \textbf{DriftQL} & \textbf{ReBRAC-v2} \\
\midrule
Bayesian-development defaults & 6 & 31.8 & 37.5 & 31.8 & \textbf{63.9} \\
Other coefficient-selection defaults & 4 & 79.1 & 72.1 & 64.3 & \textbf{98.1} \\
Tasks excluded from both selection stages & 40 & 50.0 & 52.3 & 54.2 & \textbf{74.1} \\
\midrule
Subset: categories absent from Bayesian development & 16 & 79.8 & 71.1 & 70.8 & \textbf{93.5} \\
\bottomrule
\end{tabular}
\end{adjustbox}

\end{table}

\paragraph{D4RL AntMaze and Adroit.}
Table~\ref{tab:main-results} also shows that the shared ReBRAC-v2 recipe transfers strongly to D4RL, obtaining the highest suite average in the main comparison on both AntMaze (90.2) and Adroit (33.6). On AntMaze, ReBRAC-v2 leads on three of six datasets, ties FAC on Large Diverse, and is within 3 points of the best score on each remaining dataset. On Adroit, the aggregate gain is concentrated in four tasks: ReBRAC-v2 achieves 114.9 and 110.7 on Pen Human and Pen Cloned, 24.0 on Door Cloned, and 3.6 on Relocate Cloned. It remains close to the best results on both Hammer datasets and Relocate Human, while Door Human remains effectively unsolved by all compared methods.

\paragraph{Failure Case.}
Cube Double is the clear exception: ReBRAC-v2 scores 8.3 versus ReBRAC's 12.2 and GFP's best result of 40.1, despite inclusion in shared-recipe development. Removing critic regularization or residual connections improves Cube Double while reducing average performance. This suggests a possible task-specific critic mismatch, but does not establish a mechanism; we have no conclusive explanation.

\paragraph{Beyond the Common OGBench Subset.}
Strong common-task performance motivates a broader test across the full original state-based single-task suite. The 185 additional tasks average 40.6; several stitching datasets and larger Cube tasks remain challenging despite near-perfect scores on some navigation and noisy manipulation datasets. Supplementary Tables~\ref{tab:extended-ogbench} and~\ref{tab:extended-ogbench-hyperparameters} provide per-task results and coefficients as a baseline for future methods. No compared baseline reports complete coverage, so these tasks are excluded from comparative aggregates. The shared recipe is unchanged, although some coefficient grids are extended.

\paragraph{Transfer and Adaptation Cost.}
Architecture, optimization, regularization, warm-ups, and inference remain frozen after development on six OGBench tasks. D4RL changes only the two cloning-coefficient grids, without another structural search. The recipe also transfers to four OGBench categories absent from Bayesian development, with each default-task coefficient pair reused on four other tasks. The approximately 600 Bayesian proposals are an amortized shared-recipe search, not per-task tuning or low total development compute. Main-comparison selection uses 16 pairs; released baseline configurations vary additional settings (Supplementary Table~\ref{tab:released-tuning-envelopes}). Selection uses simulator returns and is environment-assisted, although every policy update remains offline; the supplementary material details costs.

\section{Ablation Studies}
\label{sec:ablation-studies}

We first change one component at a time on six default development tasks, keeping all other hyperparameters, including both cloning coefficients, fixed as in ReBRAC \citep{tarasov2024revisiting}. Table~\ref{tab:ablations} summarizes these fixed-recipe effects; Supplementary Table~\ref{tab:ablations-full} reports task-level results. We then separately retune the remaining cloning coefficients for four objective variants and evaluate the selected settings with new seeds (Table~\ref{tab:objective-retuned-summary}). This distinguishes direct substitution from coefficient adaptation while keeping the rest of the recipe fixed.

\begin{figure}[t]
\centering
\begin{minipage}[b]{0.36\textwidth}
  \centering
  \includegraphics[width=0.88\linewidth]{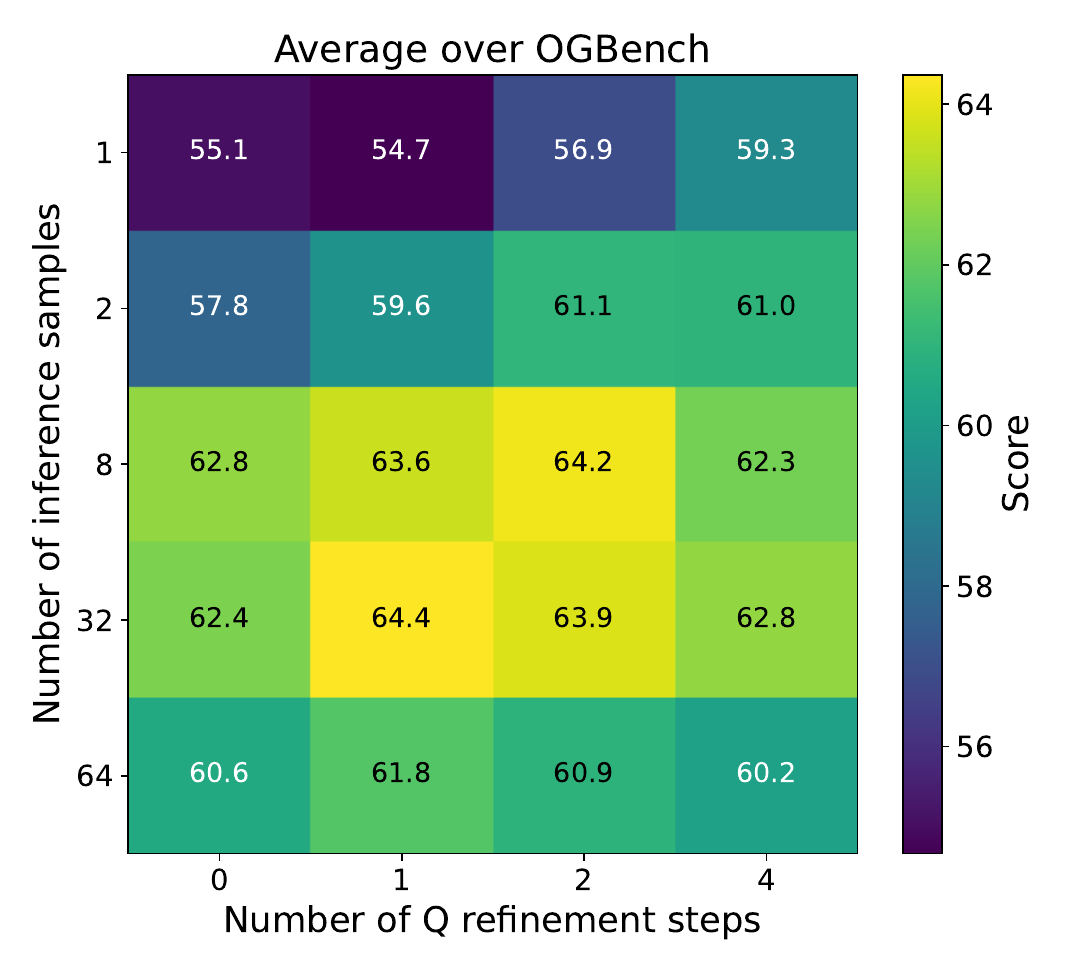}
  \par\vspace{2pt}\small (a) Inference-time sampling and Q-refinement.
\end{minipage}
\hfill
\begin{minipage}[b]{0.56\textwidth}
  \centering
  \includegraphics[width=0.88\linewidth]{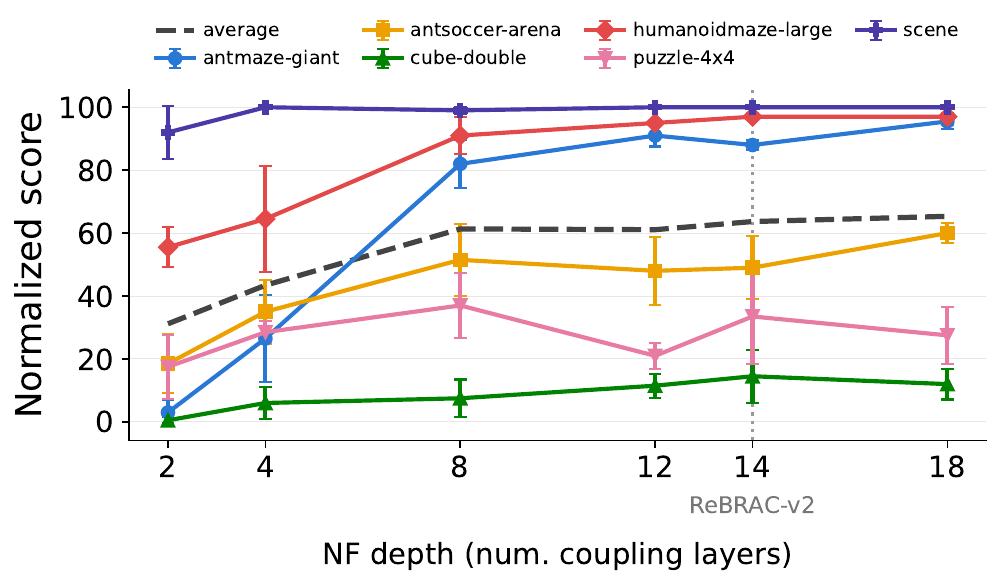}
  \par\vspace{2pt}\small (b) Normalizing-flow depth.
\end{minipage}
\caption{Inference-time compute and flow-capacity ablations. (a) Across the six OGBench development tasks, sampling provides the clearest average gain; refinement has task-dependent effects. (b) Depth is critical below eight layers, after which performance largely plateaus.}
\label{fig:inference-depth-ablations}
\label{fig:inference-ablation}
\label{fig:nf-depth-ablation}

\end{figure}

\paragraph{Objectives and Training Schedule.}
The clearest fixed-recipe sensitivity concerns the mixed cloning objective. Removing the auxiliary MSE+MAE term reduces the six-task average by 88.4\%; retaining only MAE or only MSE in the auxiliary term reduces it by 22.9\% and 18.8\%. Removing the flow likelihood term costs 15.3\%. These substitutions also change the relative strength of behavior regularization, motivating a separate coefficient-retuning experiment.

We select each variant's coefficients using four tuning seeds and evaluate them on five disjoint training seeds. Retuning raises the NLL-only average from 7.4 to 18.7, MSE+MAE without NLL from 53.9 to 57.7, NLL+MSE from 51.7 to 57.4, and NLL+MAE from 49.1 to 59.7 (Table~\ref{tab:objective-retuned-summary}). The latter is closest to the full objective's reference average of 63.7, with a remaining 6.2\% gap; the other gaps are 70.6\%, 9.3\%, and 9.8\%, respectively. Coefficient adaptation therefore explains part, but not all, of the fixed-recipe degradation. The complete objective retains the highest observed average within these searches, although this is not a claim of statistically significant separation or intrinsic superiority under joint retuning of the entire recipe. Supplementary Tables~\ref{tab:objective-retuned-full} and~\ref{tab:objective-retuned-coefficients} give task-level scores and selected coefficients.

\begin{table}[!t]
\caption{Six-task averages before and after coefficient retuning. Retuned variants use five seeds disjoint from selection; the reference is the original four-seed ablation result. $\Delta$ compares the retuned mean with the full objective. Supplementary Table~\ref{tab:objective-retuned-full} reports task-level scores.}
\label{tab:objective-retuned-summary}
\centering\small
\setlength{\tabcolsep}{4pt}
\begin{adjustbox}{max width=\linewidth}
\begin{tabular}{@{}lrrr@{}}
\toprule
BC objective & Fixed & Retuned & $\Delta$ (\%) \\
\midrule
ReBRAC-v2 (reference) & 63.7 & --- & --- \\
\midrule
NLL only & 7.4 & 18.7 & -70.6 \\
MSE + MAE & 53.9 & 57.7 & -9.3 \\
NLL + MSE & 51.7 & 57.4 & -9.8 \\
NLL + MAE & 49.1 & 59.7 & -6.2 \\
\bottomrule\end{tabular}
\end{adjustbox}

\end{table}

Staged training also contributes: removing both warm-ups lowers the average by 13.9\%, while removing only critic or behavior-cloning warm-up costs 9.6\% and 5.6\%. The larger effect of critic warm-up is consistent with the intended role of preventing a randomly initialized critic from immediately shaping the actor. This remains an empirical interpretation, but the direction is consistent across the joint and individual warm-up ablations.

With all other values fixed, minimum target aggregation reduces the average by 93.5\%, and AdamW by 65.2\%. However, Bayesian search found competitive runs with minimum aggregation and with AdamW when other hyperparameters had different values. These ablations therefore show that neither is a drop-in replacement for the final recipe, not that either is always worse.

\paragraph{Architecture and Regularization.}
Architecture and regularization choices have smaller and less uniform effects. Reducing the ensemble from four critics to two, replacing GSP with ReLU, and removing critic residual connections reduce the average by 4.7\%, 3.5\%, and 5.1\%. Removing actor regularization costs 6.5\%, while removing critic or all regularizers costs only 1.6\% or 1.0\%. The combined effect being smaller than the actor-only effect underscores interaction and estimation noise rather than a monotonic contribution from each regularizer. Exact definitions are given in the supplementary material.

Critic ablations expose task-specific trade-offs. Removing critic regularization raises AntSoccer from 49.0 to 59.5 and Cube Double from 14.5 to 24.5, but lowers AntMaze from 88.0 to 59.0, producing a 1.6\% aggregate decrease. Removing residual connections likewise improves AntSoccer and Cube Double to 51.0 and 19.5, but lowers HumanoidMaze from 97.0 to 75.5. These patterns are compatible with task-specific critic mismatch, but do not identify its mechanism. Removing next-state prediction costs 4.1\%, supporting a modest auxiliary-representation benefit.

Disabling the evaluation EMA actor improves the average by $0.7\%$, providing no controlled evidence that EMA helps and contradicting its selection by noisy one-seed development. We retain it for protocol consistency, but do not count it as an independently validated contribution.

\paragraph{Flow Depth.}
Figure~\ref{fig:inference-depth-ablations}(b) isolates flow capacity. Relative to 14 layers, depths 2 and 4 cost 51.0\% and 31.9\%; depths 8 and 12 are within 4.1\%, while depth 18 improves the average by 2.5\%. Capacity is therefore critical at shallow depths, but performance largely plateaus from 8 to 18 layers; the selected depth is competitive rather than a unique optimum. Supplementary Table~\ref{tab:nf-depth} gives task-level results.

\paragraph{Inference-Time Compute.}
In Figure~\ref{fig:inference-depth-ablations}(a), eight samples without refinement improve the six-task average by 14.1\% over one sample; four refinement steps improve the one-sample score by 7.6\%. The best setting, $(K,J)=(32,1)$, scores 64.4; our preselected $(32,2)$ scores 63.9. Refinement gains are smaller and task-dependent (Supplementary Figure~\ref{fig:inference-grid-tasks}). Applying the same grid to FAC, GFP, and DriftQL yields modest best-case gains of 1.3--3.0 points. ReBRAC-v2's worst tested aggregate (54.7) exceeds the best baseline setting (53.5); Supplementary Table~\ref{tab:inference-compute-comparison} details the comparison.

\paragraph{Inference Runtime.}
On an NVIDIA GeForce RTX 3090, a fused, device-resident implementation takes 1.59\,ms for single-state inference with $(K,J)=(32,2)$, compared with 1.11\,ms for actor-only sampling and 1.21\,ms for best-of-32 selection without refinement. These are means of per-category median latencies across the ten common OGBench categories. Full inference adds 0.47\,ms, with a mean within-category overhead ratio of $1.43\times$. The training-free benchmark uses random network parameters and excludes compilation, CPU--GPU transfers, and environment execution; it measures inference computation rather than end-to-end control-loop latency. Supplementary Table~\ref{tab:fused-inference-1} reports per-category single-state measurements.

\begin{table}[!t]
\caption{Six-task fixed-recipe substitutions without retuning. $\Delta$ measures sensitivity of the selected recipe, not independently retuned component importance; shading indicates degradation. $^\dagger$Bayesian search found competitive runs with this alternative and different surrounding hyperparameters.}
\label{tab:ablations}
\centering
\small
\setlength{\tabcolsep}{3.5pt}
\renewcommand{\arraystretch}{1.05}
\begin{adjustbox}{max width=\linewidth}
\begin{tabular}{@{}lrr@{}}
\toprule
\textbf{Ablation} & \textbf{Avg.} & $\boldsymbol{\Delta}$ (\%) \\
\midrule
ReBRAC-v2 & 63.7 & --- \\
\midrule
min training aggregation$^\dagger$ & \cellcolor{red!25}4.2 & \cellcolor{red!25}-93.5 \\
w/o auxiliary BC loss & \cellcolor{red!25}7.4 & \cellcolor{red!25}-88.4 \\
AdamW instead of Kron$^\dagger$ & \cellcolor{red!25}22.2 & \cellcolor{red!25}-65.2 \\
aux BC: MAE only & \cellcolor{red!18}49.1 & \cellcolor{red!18}-22.9 \\
aux BC: MSE only & \cellcolor{red!18}51.7 & \cellcolor{red!18}-18.8 \\
w/o NF BC loss & \cellcolor{red!14}53.9 & \cellcolor{red!14}-15.3 \\
w/o warm-up stages & \cellcolor{red!14}54.8 & \cellcolor{red!14}-13.9 \\
w/o critic warm-up & \cellcolor{red!12}57.6 & \cellcolor{red!12}-9.6 \\
w/o actor regularizers & \cellcolor{red!10}59.5 & \cellcolor{red!10}-6.5 \\
w/o BC warm-up & \cellcolor{red!10}60.1 & \cellcolor{red!10}-5.6 \\
w/o critic residuals & \cellcolor{red!10}60.4 & \cellcolor{red!10}-5.1 \\
2 critics & \cellcolor{red!10}60.7 & \cellcolor{red!10}-4.7 \\
w/o next-state prediction & \cellcolor{red!10}61.1 & \cellcolor{red!10}-4.1 \\
ReLU activation & \cellcolor{red!8}61.4 & \cellcolor{red!8}-3.5 \\
w/o critic regularizers & \cellcolor{red!6}62.7 & \cellcolor{red!6}-1.6 \\
w/o all regularizers & \cellcolor{red!6}63.0 & \cellcolor{red!6}-1.0 \\
w/o actor EMA & \cellcolor{green!12}64.1 & \cellcolor{green!12}+0.7 \\
\bottomrule
\end{tabular}
\end{adjustbox}

\end{table}

\section{Limitations and Future Work}

The shared recipe is selected from approximately 600 noisy one-seed proposals on six tasks drawn from the final OGBench families. Transfer to their remaining tasks and to D4RL provides stronger evidence, but does not eliminate post-selection optimism. Selection uses environment returns, and the total development cost is substantial despite the narrow final adaptation interface. Fixed-recipe ablations cannot fully separate interactions. Our coefficient-retuned objective experiments address one part of this uncertainty, but their search budgets differ with the number of remaining coefficients, and structural and optimization choices remain frozen. Alternatives may recover further after joint retuning; multi-seed shared-recipe development would also strengthen attribution. Evaluation is limited to state-based, fully offline simulation. Pixel observations, offline-to-online tuning, and real-robot transfer remain open, as do improved critics for Cube Double and application of this modernization recipe beyond ReBRAC.

\section{Conclusion}

ReBRAC-v2 shows that systematic modernization of a conventional offline actor-critic can yield the strongest compared OGBench and D4RL aggregates with a shared recipe and 16-point coefficient adaptation. Evaluation on all 235 original state-based single-task OGBench problems provides a reference baseline for future research, including settings that remain poorly solved.

\clearpage
\bibliography{references}
\bibliographystyle{rlj}

\clearpage
\begin{center}
\textbf{\LARGE Supplementary Materials}
\end{center}
\raggedbottom
\appendix
\FloatBarrier
\section{Method and Algorithm Details}
\label{sec:method-details}

\paragraph{Implementation provenance.}
Our implementation extends the CORL offline-RL codebase \citep{tarasov2024corl}. The ReBRAC-v2 components and configurations described in this paper are implemented on top of that foundation.

\paragraph{Actor and critic architecture.}
Each actor block applies a state-conditioned RealNVP coupling transform \citep{dinh2016density} followed by a learned $W=PLU$ linear flow adapted from invertible $1\times1$ convolutions \citep{kingma2018glow}. PLU mixes action dimensions while retaining efficient inversion and log-determinant computation. The selected actor has 14 blocks; each conditioner is a two-hidden-layer MLP of width 256 with GSP, LayerNorm \citep{ba2016layer}, and dropout. A final $\tanh$ bounds actions. The selected critic has a five-layer residual trunk with LayerNorm and dropout in each residual branch, followed by separate two-layer value and next-state heads. GSP is
\begin{equation}
\sigma(x)=\operatorname{GELU}(x)\left(1+0.5\operatorname{sinc}(x)\right).
\end{equation}
The search selects GSP over ReLU and SiLU, while residual connections remain enabled in all search configurations.

\paragraph{Numerical parameterization.}
Let $G_{\min}$ and $G_{\max}$ be the minimum and maximum Monte Carlo returns in the offline dataset and $\Delta=G_{\max}-G_{\min}$. The 201 critic bins partition $[G_{\min}-0.025\Delta,G_{\max}+0.025\Delta]$. The HL-Gauss projection integrates a Gaussian with standard deviation $0.75$ bin widths over each bin and renormalizes its mass within this support, using $10^{-6}$ for numerical stability; out-of-support targets are not hard-clipped. The next-state MSE has coefficient 1 and uses the raw, unnormalized state vectors because state normalization is disabled in all reported configurations.

The selected ``uniform'' flow base is implemented by sampling $u\sim\mathcal U([-1,1]^{d_{\mathcal A}})$ and setting $z=\operatorname{atanh}(u)$ before the coupling flow; likelihood evaluation includes the corresponding transformed-uniform density. Before applying the inverse final $\tanh$, dataset actions are clipped coordinatewise to $[-1+10^{-6},1-10^{-6}]$, and the log-Jacobian uses the same $10^{-6}$ stabilizer. This defines finite likelihoods for dataset actions at the action bounds.

\paragraph{Auxiliary prediction and EMA.}
The transition head minimizes $\mathcal L_{\mathrm{dyn}}=\|g_\psi(h_\theta(s,a))-s'\|_2^2$ for every critic update. Unlike transition-prediction pretraining that discards the objective during offline RL fine-tuning \citep{park2025pretraining}, ReBRAC-v2 retains it throughout. Actor averaging uses
\begin{equation}
\phi_{\mathrm{EMA}}\leftarrow(1-\tau_{\mathrm{EMA}})\phi_{\mathrm{EMA}}+\tau_{\mathrm{EMA}}\phi,
\qquad \tau_{\mathrm{EMA}}=0.005,
\end{equation}
only for evaluation. Critic targets use the current actor rather than a target or EMA actor.

\begin{algorithm}[H]
\caption{ReBRAC-v2 training and evaluation}
\label{alg:rebrac-v2}
\begin{algorithmic}[1]
\STATE Initialize flow actor $\pi_\phi$, EMA actor, four critics, and target critics
\FOR{$t=1,\ldots,T$}
  \STATE Sample $(s,a,r,s',d)\sim\mathcal D$
  \IF{$t\leq T_{\mathrm{IL}}$}
    \STATE Update actor with $\mathcal L_{\mathrm{BC}}$ and update its EMA
  \ELSE
    \STATE Sample noisy $a'\sim\pi_\phi(\cdot\mid s')$ and form the mean-ensemble Bellman target
    \STATE Update all critics with categorical Bellman loss plus $\mathcal L_{\mathrm{dyn}}$
    \IF{$t>T_{\mathrm{IL}}+T_Q$ and $t\bmod d=0$}
      \STATE Update actor with $-\lambda Q_{\min}(s,\tilde a)+\mathcal L_{\mathrm{BC}}$ and update its EMA
    \ENDIF
    \STATE Polyak-update target critics on schedule
  \ENDIF
\ENDFOR
\STATE At evaluation, sample $K$ EMA-actor actions, apply $J$ normalized $Q_{\min}$-gradient steps, and execute the candidate maximizing $Q_{\min}$
\end{algorithmic}
\end{algorithm}

\FloatBarrier
\subsection{Explored Variants That Did Not Improve Results}
\label{sec:negative-results}

We also evaluated several plausible extensions that were not retained because they did not improve the selected recipe within our development budget. These observations concern the tested implementations and hyperparameter ranges; they should not be read as general negative results for the underlying ideas.

\paragraph{Action chunking and multi-step targets.}
We chunked both the policy output and critic input following Q-chunking, which performs RL directly in a temporally extended action space \citep{li2025qchunking}. We also integrated the strong \emph{NS} baseline studied in Decoupled Q-Chunking \citep{li2025decoupled}: the policy remains single-step and a one-action critic is trained with an uncorrected $n$-step return target. Neither Q-chunking nor NS improved our development results over the selected single-action policy and one-step Bellman backup, so we retained the simpler formulation.

\paragraph{Previous-transition context.}
We separately augmented the actor and critic inputs with the previous state only, the previous action only, or both, aiming to provide short-term context beyond the current transition. All three variants harmed performance in our setup, so the selected method retains the standard state input for the actor and state-action input for the critic.

\paragraph{More expressive flow transforms.}
We replaced affine coupling transforms with Neural Spline Flows, whose monotonic rational-quadratic splines provide a more flexible invertible elementwise transformation while preserving exact likelihood and analytic inversion \citep{durkan2019neural}. Surprisingly, this additional expressivity did not improve performance even after substantial tuning of the spline-specific hyperparameters. We therefore retained affine coupling, which is also simpler and cheaper.

\paragraph{Multi-sample Bellman targets.}
We tested drawing multiple next actions from the target policy, evaluating each with the target critics, and combining the resulting values using minimum, mean, or maximum aggregation. This was intended to make the Bellman target less dependent on one policy sample or to favor higher-value candidates. The Bayesian search nevertheless preferred a single sampled action, and the multi-sample variants provided no consistent benefit. This differs from evaluation-time candidate selection: multi-sample inference improves the executed action without changing the critic's training target.

\FloatBarrier
\section{Full D4RL Results}

\begin{table}[!htbp]
\caption{Task-level D4RL results underlying Table~\ref{tab:main-results}. ReBRAC-v2, FAC, GFP, and DriftQL entries are our reruns; uncertainty values and evaluation details are described in Supplementary Section~\ref{sec:baseline-protocols}.}
\label{tab:d4rl-full}
\centering
\appendixtableformat
\begin{adjustbox}{max width=\linewidth}
\begin{tabular}{@{}lrrrrrrrrr@{}}
\toprule
\textbf{Dataset} & \textbf{BC} & \textbf{IQL} & \textbf{ReBRAC} & \textbf{IFQL} & \textbf{FQL} & \textbf{FAC} & \textbf{GFP} & \textbf{DriftQL} & \textbf{ReBRAC-v2} \\
\midrule
antmaze-umaze-v2 & 55 & 77 & 98 & 92 & 96 & 97.3 & \underline{98.3} & 95.1 & \textbf{99.3} \\
antmaze-umaze-diverse-v2 & 47 & 54 & 84 & 62 & 89 & \textbf{92.7} & 87.9 & 87.6 & \underline{91.7} \\
antmaze-medium-play-v2 & 0 & 66 & \underline{90} & 56 & 78 & 79.7 & 82.9 & 75.7 & \textbf{90.8} \\
antmaze-medium-diverse-v2 & 1 & 74 & \underline{84} & 60 & 71 & 69.7 & 62.9 & 73.0 & \textbf{89.5} \\
antmaze-large-play-v2 & 0 & 42 & 52 & 55 & \underline{84} & \textbf{86.2} & 82.9 & 78.2 & 83.4 \\
antmaze-large-diverse-v2 & 0 & 30 & 64 & 64 & 83 & \textbf{86.5} & \underline{84.9} & 54.0 & \textbf{86.5} \\
\midrule
pen-human-v1 & 71 & 78 & \underline{103} & 71 & 53 & 60.5 & 77.0 & 44.7 & \textbf{114.9} \\
pen-cloned-v1 & 52 & 83 & \underline{103} & 80 & 74 & 95.1 & 81.0 & 53.6 & \textbf{110.7} \\
door-human-v1 & 2 & \underline{3} & 0 & \textbf{7} & 0 & 2.6 & 0.1 & -0.1 & 0.0 \\
door-cloned-v1 & 0 & 3 & 0 & 2 & 2 & \underline{4.9} & 0.5 & 0.0 & \textbf{24.0} \\
hammer-human-v1 & 3 & 2 & 0 & 3 & 1 & \textbf{4.3} & 1.6 & 0.4 & \underline{4.2} \\
hammer-cloned-v1 & 1 & 2 & 5 & 2 & \underline{11} & 7.7 & 9.4 & 1.0 & \textbf{11.1} \\
relocate-human-v1 & 0 & 0 & 0 & 0 & 0 & \underline{0.1} & \textbf{0.3} & -0.1 & \textbf{0.3} \\
relocate-cloned-v1 & 0 & 0 & \underline{2} & 0 & 0 & 0.2 & 1.5 & 0.1 & \textbf{3.6} \\
\bottomrule
\end{tabular}
\end{adjustbox}

\end{table}

\FloatBarrier
\section{Full Ablation Results}

\begin{table}[p]
\caption{Task-level one-factor ablations without hyperparameter retuning. Entries are mean $\pm$ standard deviation over seeds; $\Delta$ is the relative six-task average change. Cells are shaded relative to ReBRAC-v2. $^\dagger$Bayesian search found competitive runs with this alternative when other hyperparameters had different values.}
\label{tab:ablations-full}
\centering
\footnotesize
\setlength{\tabcolsep}{3pt}
\renewcommand{\arraystretch}{1.08}
\begin{adjustbox}{max width=\linewidth}
\begin{tabular}{@{}lcccrr@{}}
\toprule
Ablation & \shortstack{AntMaze\\Giant} & \shortstack{AntSoccer\\Arena} & \shortstack{Cube\\Double} & Avg. & $\Delta$ (\%) \\
\midrule
ReBRAC-v2 & 88.0 $\pm$ 1.6 & 49.0 $\pm$ 10.0 & 14.5 $\pm$ 8.4 & 63.7 & --- \\
\midrule
min training aggregation$^\dagger$ & \cellcolor{red!12}9.0 $\pm$ 8.9 & \cellcolor{red!12}0.0 $\pm$ 0.0 & \cellcolor{red!12}0.0 $\pm$ 0.0 & 4.2 & -93.5 \\
w/o auxiliary BC loss & \cellcolor{red!12}0.0 $\pm$ 0.0 & \cellcolor{red!12}0.5 $\pm$ 1.0 & \cellcolor{red!12}0.0 $\pm$ 0.0 & 7.4 & -88.4 \\
AdamW instead of Kron$^\dagger$ & \cellcolor{red!12}15.0 $\pm$ 27.4 & \cellcolor{red!12}6.5 $\pm$ 4.1 & \cellcolor{red!12}3.5 $\pm$ 3.4 & 22.2 & -65.2 \\
aux BC: MAE only & \cellcolor{green!12}89.0 $\pm$ 2.6 & \cellcolor{red!12}40.0 $\pm$ 15.9 & \cellcolor{red!12}5.5 $\pm$ 1.0 & 49.1 & -22.9 \\
aux BC: MSE only & 88.0 $\pm$ 1.6 & \cellcolor{red!12}37.5 $\pm$ 9.8 & \cellcolor{red!12}3.5 $\pm$ 3.0 & 51.7 & -18.8 \\
w/o NF BC loss & \cellcolor{green!12}93.5 $\pm$ 5.7 & \cellcolor{red!12}26.5 $\pm$ 6.0 & \cellcolor{red!12}2.0 $\pm$ 2.8 & 53.9 & -15.3 \\
w/o warm-up stages & \cellcolor{green!12}93.0 $\pm$ 2.6 & \cellcolor{red!12}44.0 $\pm$ 1.6 & \cellcolor{red!12}3.0 $\pm$ 3.5 & 54.8 & -13.9 \\
w/o critic warm-up & \cellcolor{green!12}94.0 $\pm$ 5.4 & \cellcolor{red!12}45.5 $\pm$ 3.4 & \cellcolor{red!12}11.0 $\pm$ 7.0 & 57.6 & -9.6 \\
w/o actor regularizers & \cellcolor{green!12}92.5 $\pm$ 3.4 & \cellcolor{red!12}36.0 $\pm$ 5.9 & \cellcolor{red!12}8.5 $\pm$ 5.7 & 59.5 & -6.5 \\
w/o BC warm-up & \cellcolor{green!12}90.0 $\pm$ 4.0 & \cellcolor{red!12}39.5 $\pm$ 8.2 & \cellcolor{red!12}10.5 $\pm$ 10.1 & 60.1 & -5.6 \\
w/o critic residuals & \cellcolor{red!12}84.0 $\pm$ 4.3 & \cellcolor{green!12}51.0 $\pm$ 10.1 & \cellcolor{green!12}19.5 $\pm$ 10.8 & 60.4 & -5.1 \\
2 critics & \cellcolor{red!12}87.5 $\pm$ 1.0 & \cellcolor{red!12}46.0 $\pm$ 13.4 & \cellcolor{red!12}6.0 $\pm$ 3.7 & 60.7 & -4.7 \\
w/o next-state prediction & \cellcolor{red!12}83.5 $\pm$ 5.7 & \cellcolor{red!12}38.5 $\pm$ 5.3 & \cellcolor{red!12}9.0 $\pm$ 5.8 & 61.1 & -4.1 \\
ReLU activation & \cellcolor{green!12}89.5 $\pm$ 1.9 & 49.0 $\pm$ 4.8 & \cellcolor{red!12}9.5 $\pm$ 4.4 & 61.4 & -3.5 \\
w/o critic regularizers & \cellcolor{red!12}59.0 $\pm$ 19.9 & \cellcolor{green!12}59.5 $\pm$ 14.3 & \cellcolor{green!12}24.5 $\pm$ 8.1 & 62.7 & -1.6 \\
w/o all regularizers & \cellcolor{red!12}85.0 $\pm$ 10.1 & \cellcolor{green!12}53.5 $\pm$ 6.6 & \cellcolor{red!12}14.0 $\pm$ 13.0 & 63.0 & -1.0 \\
w/o actor EMA & \cellcolor{green!12}95.0 $\pm$ 3.8 & \cellcolor{red!12}48.5 $\pm$ 4.4 & \cellcolor{red!12}12.5 $\pm$ 5.3 & 64.1 & +0.7 \\
\bottomrule
\end{tabular}
\end{adjustbox}
\par\medskip
\begin{adjustbox}{max width=\linewidth}
\begin{tabular}{@{}lcccrr@{}}
\toprule
Ablation & \shortstack{HumanoidMaze\\Large} & \shortstack{Puzzle\\4x4} & Scene & Avg. & $\Delta$ (\%) \\
\midrule
ReBRAC-v2 & 97.0 $\pm$ 1.2 & 33.5 $\pm$ 15.0 & 100.0 $\pm$ 0.0 & 63.7 & --- \\
\midrule
min training aggregation$^\dagger$ & \cellcolor{red!12}1.5 $\pm$ 1.9 & \cellcolor{red!12}12.0 $\pm$ 3.7 & \cellcolor{red!12}2.5 $\pm$ 1.9 & 4.2 & -93.5 \\
w/o auxiliary BC loss & \cellcolor{red!12}0.0 $\pm$ 0.0 & \cellcolor{red!12}0.0 $\pm$ 0.0 & \cellcolor{red!12}44.0 $\pm$ 29.6 & 7.4 & -88.4 \\
AdamW instead of Kron$^\dagger$ & \cellcolor{red!12}0.0 $\pm$ 0.0 & \cellcolor{red!12}8.5 $\pm$ 7.7 & \cellcolor{red!12}99.5 $\pm$ 1.0 & 22.2 & -65.2 \\
aux BC: MAE only & \cellcolor{red!12}38.0 $\pm$ 32.0 & \cellcolor{red!12}22.0 $\pm$ 2.8 & 100.0 $\pm$ 0.0 & 49.1 & -22.9 \\
aux BC: MSE only & \cellcolor{red!12}82.5 $\pm$ 9.1 & \cellcolor{red!12}2.0 $\pm$ 2.8 & \cellcolor{red!12}96.5 $\pm$ 1.0 & 51.7 & -18.8 \\
w/o NF BC loss & \cellcolor{red!12}88.0 $\pm$ 5.2 & 33.5 $\pm$ 23.5 & \cellcolor{red!12}80.0 $\pm$ 11.5 & 53.9 & -15.3 \\
w/o warm-up stages & \cellcolor{red!12}88.5 $\pm$ 9.0 & \cellcolor{red!12}0.5 $\pm$ 1.0 & 100.0 $\pm$ 0.0 & 54.8 & -13.9 \\
w/o critic warm-up & \cellcolor{red!12}88.0 $\pm$ 14.9 & \cellcolor{red!12}7.0 $\pm$ 5.8 & 100.0 $\pm$ 0.0 & 57.6 & -9.6 \\
w/o actor regularizers & \cellcolor{red!12}92.5 $\pm$ 1.9 & \cellcolor{red!12}31.0 $\pm$ 11.9 & \cellcolor{red!12}96.5 $\pm$ 3.4 & 59.5 & -6.5 \\
w/o BC warm-up & \cellcolor{red!12}94.5 $\pm$ 1.9 & \cellcolor{red!12}26.5 $\pm$ 8.1 & \cellcolor{red!12}99.5 $\pm$ 1.0 & 60.1 & -5.6 \\
w/o critic residuals & \cellcolor{red!12}75.5 $\pm$ 12.8 & \cellcolor{red!12}32.5 $\pm$ 6.6 & 100.0 $\pm$ 0.0 & 60.4 & -5.1 \\
2 critics & \cellcolor{red!12}94.5 $\pm$ 1.9 & \cellcolor{red!12}30.0 $\pm$ 14.3 & 100.0 $\pm$ 0.0 & 60.7 & -4.7 \\
w/o next-state prediction & \cellcolor{green!12}98.0 $\pm$ 0.0 & \cellcolor{green!12}37.5 $\pm$ 9.3 & 100.0 $\pm$ 0.0 & 61.1 & -4.1 \\
ReLU activation & \cellcolor{red!12}91.5 $\pm$ 1.0 & \cellcolor{red!12}29.0 $\pm$ 9.9 & 100.0 $\pm$ 0.0 & 61.4 & -3.5 \\
w/o critic regularizers & 97.0 $\pm$ 2.6 & \cellcolor{green!12}36.0 $\pm$ 13.5 & 100.0 $\pm$ 0.0 & 62.7 & -1.6 \\
w/o all regularizers & \cellcolor{green!12}98.5 $\pm$ 1.9 & \cellcolor{red!12}28.5 $\pm$ 10.1 & \cellcolor{red!12}98.5 $\pm$ 1.9 & 63.0 & -1.0 \\
w/o actor EMA & \cellcolor{red!12}95.0 $\pm$ 1.2 & \cellcolor{green!12}34.0 $\pm$ 10.8 & \cellcolor{red!12}99.5 $\pm$ 1.0 & 64.1 & +0.7 \\
\bottomrule
\end{tabular}
\end{adjustbox}
\end{table}

\FloatBarrier
\subsection{Coefficient-Retuned Objective Ablations}
\label{sec:objective-retuning}

The fixed-recipe comparisons change the actor objective without adapting its coefficients. To test whether this explains their degradation, we retune each of four variants independently on the same six default development tasks: NLL only, MSE+MAE without NLL, NLL+MSE, and NLL+MAE. Every other setting remains fixed, including the architecture, optimizer, warm-ups, regularizers, and inference procedure. All runs train for one million updates and evaluate the final checkpoint using 50 episodes.

\paragraph{Selection and confirmation.}
For NLL only, the auxiliary coefficient is zero and the NLL coefficient is selected from $\{10^{-5},3\times10^{-5},10^{-4},3\times10^{-4},10^{-3},3\times10^{-3},10^{-2},3\times10^{-2}\}$. For MSE+MAE without NLL, the NLL coefficient is zero and the auxiliary coefficient is selected from $\{0.003,0.01,0.03,0.1,0.3,0.5,1,3\}$. For each of NLL+MSE and NLL+MAE, we search the Cartesian product $\alpha_{\mathrm{NF}}\in\{0.0001,0.001,0.003,0.01\}$ and $\alpha_{\mathrm{aux}}\in\{0.03,0.1,0.5,1\}$. The two-coefficient searches therefore use the same 16-pair grid as the full objective, whereas the single-coefficient variants use eight settings each.

Selection maximizes the mean final-checkpoint score across training seeds 0--3, separately for each variant and task. These grids comprise 1,152 training runs. Exact ties are resolved by the smaller auxiliary coefficient, then the smaller NLL coefficient. In particular, the NLL-only AntMaze and HumanoidMaze candidates tie at zero, so their selected coefficients should not be interpreted as uniquely preferable. We freeze the selected settings and perform 120 confirmation runs: four variants, six tasks, and five training seeds 10--14. Table~\ref{tab:objective-retuned-coefficients} lists the selected pairs. Evaluation-environment seed 42 is unchanged.

\begin{table}[!htbp]
\caption{Coefficient-retuned objective ablations on the six default development tasks. New rows report mean $\pm$ sample standard deviation across training seeds 10--14; the reference repeats the original four-seed result from Table~\ref{tab:ablations-full}, not a new run. All scores use the final checkpoint after 1M updates and 50 evaluation episodes. Shading and relative average changes compare against that reference; averages weight tasks equally.}
\label{tab:objective-retuned-full}
\centering
\footnotesize
\setlength{\tabcolsep}{2pt}\appendixtableformat
\resizebox{\linewidth}{!}{\begin{tabular}{@{}lcccccc|rr@{}}
\toprule
BC objective & \shortstack{AntMaze\\Giant} & \shortstack{AntSoccer\\Arena} & \shortstack{Cube\\Double} & \shortstack{HumanoidMaze\\Large} & \shortstack{Puzzle\\4x4} & Scene & Avg. & $\Delta$ (\%) \\
\midrule
ReBRAC-v2 & 88.0 $\pm$ 1.6 & 49.0 $\pm$ 10.0 & 14.5 $\pm$ 8.4 & 97.0 $\pm$ 1.2 & 33.5 $\pm$ 15.0 & 100.0 $\pm$ 0.0 & 63.7 & --- \\
\midrule
NLL only & \cellcolor{red!12}0.0 $\pm$ 0.0 & \cellcolor{red!12}0.4 $\pm$ 0.9 & \cellcolor{red!12}0.8 $\pm$ 1.1 & \cellcolor{red!12}0.0 $\pm$ 0.0 & \cellcolor{red!12}25.2 $\pm$ 22.5 & \cellcolor{red!12}86.0 $\pm$ 15.9 & 18.7 & -70.6 \\
MSE + MAE & \cellcolor{green!12}97.2 $\pm$ 1.8 & \cellcolor{red!12}38.4 $\pm$ 7.1 & \cellcolor{red!12}0.8 $\pm$ 1.8 & \cellcolor{red!12}94.8 $\pm$ 2.7 & \cellcolor{red!12}22.8 $\pm$ 15.5 & \cellcolor{red!12}92.4 $\pm$ 5.9 & 57.7 & -9.3 \\
NLL + MSE & \cellcolor{red!12}84.8 $\pm$ 10.9 & \cellcolor{red!12}45.6 $\pm$ 9.9 & \cellcolor{red!12}5.6 $\pm$ 4.3 & \cellcolor{red!12}91.6 $\pm$ 6.2 & \cellcolor{red!12}18.4 $\pm$ 10.9 & \cellcolor{red!12}98.4 $\pm$ 2.6 & 57.4 & -9.8 \\
NLL + MAE & \cellcolor{green!12}91.6 $\pm$ 3.3 & \cellcolor{red!12}48.0 $\pm$ 5.1 & \cellcolor{red!12}7.6 $\pm$ 6.1 & \cellcolor{red!12}90.8 $\pm$ 2.3 & \cellcolor{red!12}21.2 $\pm$ 31.8 & \cellcolor{red!12}99.2 $\pm$ 1.8 & 59.7 & -6.2 \\
\bottomrule\end{tabular}}

\end{table}

\paragraph{Interpretation.}
Retuning improves all four aggregate point estimates relative to their fixed-coefficient substitutions, most substantially for NLL only (7.4 to 18.7) and NLL+MAE (49.1 to 59.7). For NLL+MAE, the HumanoidMaze score rises from 38.0 to 90.8, showing that much of the original task-level drop can be recovered by coefficient adaptation. Nevertheless, the full objective retains the highest six-task reference average, 63.7, compared with 59.7 for NLL+MAE, 57.7 for MSE+MAE without NLL, 57.4 for NLL+MSE, and 18.7 for NLL only. The remaining relative gaps are 6.2\%, 9.3\%, 9.8\%, and 70.6\%, respectively. The ordering is not universal across tasks: MSE+MAE without NLL exceeds the reference on AntMaze (97.2 versus 88.0), as does NLL+MAE (91.6). Conversely, NLL-only remains unsuccessful on AntMaze and HumanoidMaze despite its gains on Scene and Puzzle.

These results strengthen the empirical case for the complete objective within the tested coefficient ranges, while qualifying the larger effects seen in fixed-recipe substitutions. The full-method reference is reused from the original four-seed ablation experiment, whereas the new variants use five seeds; we do not treat the resulting differences as formal significance tests. Moreover, one- and two-coefficient variants receive different numbers of candidates, and no variant receives joint architecture or optimizer retuning. This experiment therefore does not establish an intrinsic advantage of combining all three losses under unrestricted tuning.

\begin{table}[!htbp]
\caption{Selected $(\alpha_{\mathrm{NF}},\alpha_{\mathrm{aux}})$ pairs for the retuned objective ablations. Selection uses seeds 0--3; the coefficients are frozen for confirmation on seeds 10--14.}
\label{tab:objective-retuned-coefficients}
\centering\appendixtableformat
\setlength{\tabcolsep}{5pt}\begin{adjustbox}{max width=\linewidth}
\begin{tabular}{@{}lcccc@{}}
\toprule
Task & NLL only & MSE + MAE & NLL + MSE & NLL + MAE \\
\midrule
AntMaze Giant & $(10^{-5}, 0)$ & $(0, 0.01)$ & $(0.0001, 0.03)$ & $(0.0001, 0.03)$ \\
AntSoccer Arena & $(0.003, 0)$ & $(0, 0.03)$ & $(0.001, 0.1)$ & $(0.0001, 0.03)$ \\
Cube Double & $(0.01, 0)$ & $(0, 0.03)$ & $(0.003, 1)$ & $(0.001, 0.5)$ \\
HumanoidMaze Large & $(3\times10^{-5}, 0)$ & $(0, 0.03)$ & $(0.0001, 0.1)$ & $(0.0001, 0.1)$ \\
Puzzle 4x4 & $(0.003, 0)$ & $(0, 0.1)$ & $(0.001, 0.1)$ & $(0.003, 0.03)$ \\
Scene & $(0.03, 0)$ & $(0, 0.1)$ & $(0.003, 1)$ & $(0.001, 1)$ \\
\bottomrule\end{tabular}
\end{adjustbox}

\end{table}

\FloatBarrier
\section{Hyperparameter Development Details}

\paragraph{Shared-Recipe Search.}
The six development datasets are \path{antmaze-giant-navigate-singletask-v0}, \path{antsoccer-arena-navigate-singletask-v0}, \path{cube-double-play-singletask-v0}, \path{humanoidmaze-large-navigate-singletask-v0}, \path{puzzle-4x4-play-singletask-v0}, and \path{scene-play-singletask-v0}. One proposal comprises six training runs with seed 0, and its objective is their unweighted mean normalized score. We evaluate 576 proposals, or 3,456 single-task runs. Preliminary experiments fix the MSE+MAE auxiliary loss, the uniform flow base, $(K,J,\eta)=(32,2,0.01)$, and the initial coefficients $(\alpha_{\mathrm{aux}},\alpha_{\mathrm{NF}})=(0.03,3\times10^{-4})$. We subsequently freeze every architectural, optimization, regularization, training, and inference choice.

Table~\ref{tab:bayes-search-space} reports the complete search range for each parameter. Log-uniform intervals sample the parameter uniformly in log space. For \texttt{actor\_ema\_tau}, a value of 1 disables averaging. Values selected extremely close to zero, as well as regularizers rejected in preliminary experiments, are set exactly to zero in the final configuration. Table~\ref{tab:shared-hyperparameters} gives the resulting shared values.

\begingroup
\small
\setlength{\tabcolsep}{2pt}
\renewcommand{\arraystretch}{1.08}
\begin{longtable}{@{}lp{0.48\linewidth}@{}}
\caption{Hyperparameter ranges used for shared-recipe Bayesian development.}\label{tab:bayes-search-space}\\
\toprule
\textbf{Parameter} & \textbf{Search domain} \\
\midrule
\endfirsthead
\multicolumn{2}{l}{\tablename\ \thetable{} (continued)}\\
\toprule
\textbf{Parameter} & \textbf{Search domain} \\
\midrule
\endhead
\midrule
\multicolumn{2}{r}{Continued on next page}\\
\endfoot
\bottomrule
\endlastfoot
\texttt{actor\_bc\_noise} & log-uniform $[10^{-4},10^{-1}]$ \\
\texttt{activation} & \{GSP, ReLU, SiLU\} \\
\texttt{actor\_ema\_tau} & $\{0.003,0.005,0.01,1\}$ \\
\texttt{actor\_grad\_noise} & log-uniform $[10^{-5},10^{-1}]$ \\
\texttt{actor\_learning\_rate} & log-uniform $[10^{-5},3\!\times\!10^{-3}]$ \\
\texttt{actor\_wd} & log-uniform $[10^{-6},10^{-2}]$ \\
\texttt{batch\_size} & $\{512,1024,2048\}$ \\
\texttt{critic\_dropout} & uniform $[0,0.1]$ \\
\texttt{critic\_grad\_noise} & log-uniform $[10^{-8},10^{-3}]$ \\
\texttt{critic\_learning\_rate} & log-uniform $[10^{-5},3\!\times\!10^{-3}]$ \\
\texttt{critic\_n\_hiddens} & $\{4,5,6\}$ \\
\texttt{next-state objective} & \{disabled, enabled\} \\
\texttt{policy\_noise} & $\{0,0.2\}$ \\
\texttt{tau} & log-uniform $[0.005,0.05]$ \\
\texttt{use\_target\_actor} & $\{\mathrm{false},\mathrm{true}\}$ \\
\texttt{critic\_objective\_noise} & log-uniform $[10^{-6},10^{-2}]$ \\
\texttt{critic\_warmup\_epochs} & $\{0,50,100,200\}\!\times\!10^3$ updates \\
\texttt{critic\_wd} & log-uniform $[10^{-8},10^{-2}]$ \\
\texttt{il\_warmup\_epochs} & $\{0,50,100,200\}\!\times\!10^3$ updates \\
\texttt{n\_classes} & $\{51,101,201\}$ \\
\texttt{nf\_dropout} & uniform $[0.01,0.2]$ \\
\texttt{nf\_hidden\_dim} & $\{128,256,512\}$ \\
\texttt{nf\_n\_hiddens} & $\{2,3\}$ \\
\texttt{nf\_num\_layers} & $\{6,8,10,12,14,16\}$ \\
\texttt{nf\_use\_plu} & $\{\mathrm{false},\mathrm{true}\}$ \\
\texttt{num\_critics} & $\{2,4\}$ \\
\texttt{optimizer\_type} & \{Kron, Adan, AdamW\} \\
\texttt{target\_critic\_aggregation} & \{minimum, mean, maximum\} \\
\texttt{gamma} & $\{0.99,0.995,0.999\}$ \\
\end{longtable}
\endgroup

\begingroup
\small
\setlength{\tabcolsep}{2pt}
\renewcommand{\arraystretch}{1.08}
\begin{longtable}{@{}lp{0.48\linewidth}@{}}
\caption{Shared hyperparameters selected by Bayesian development and then frozen for all reported OGBench and D4RL runs. Behavior-regularization coefficients are reported separately in Table~\ref{tab:bc-coefficients}.}\label{tab:shared-hyperparameters}\\
\toprule
\textbf{Parameter} & \textbf{Selected value} \\
\midrule
\endfirsthead
\multicolumn{2}{l}{\tablename\ \thetable{} (continued)}\\
\toprule
\textbf{Parameter} & \textbf{Selected value} \\
\midrule
\endhead
\midrule
\multicolumn{2}{r}{Continued on next page}\\
\endfoot
\bottomrule
\endlastfoot
Activation & GSP \\
Actor BC noise & $1.153\!\times\!10^{-3}$ \\
Actor EMA $\tau$ & $0.005$ \\
Actor gradient noise & $2.592\!\times\!10^{-4}$ \\
Actor learning rate & $4.127\!\times\!10^{-5}$ \\
Actor weight decay & $9.632\!\times\!10^{-6}$ \\
Batch size & $512$ \\
Critic dropout & $0.004813$ \\
Critic gradient noise & $5.037\!\times\!10^{-6}$ \\
Critic learning rate & $2.393\!\times\!10^{-5}$ \\
Critic hidden layers & $5$ \\
Next-state objective & enabled \\
Target policy noise & $0.2$ \\
Target-network $\tau$ & $0.007311$ \\
Target actor & disabled \\
Critic objective noise & $0$ \\
Critic warm-up & $200{,}000$ updates \\
Critic weight decay & $3.685\!\times\!10^{-5}$ \\
IL warm-up & $100{,}000$ updates \\
Value classes & $201$ \\
Flow dropout & $0.04812$ \\
Flow hidden width & $256$ \\
Flow conditioner depth & $2$ \\
Flow coupling layers & $14$ \\
PLU mixing & enabled \\
Number of critics & $4$ \\
Optimizer & Kron \\
Target critic aggregation & mean \\
Discount $\gamma$ & $0.999$ \\
\end{longtable}
\endgroup

\paragraph{Behavior-Regularization Adaptation.}
For the main comparison, after fixing the shared recipe, we tune only $(\alpha_{\mathrm{NF}},\alpha_{\mathrm{aux}})$ using seeds 0--3. For OGBench, we evaluate
$\alpha_{\mathrm{NF}}\in\{10^{-4},10^{-3},3\times10^{-3},10^{-2}\}$ and
$\alpha_{\mathrm{aux}}\in\{0.03,0.1,0.5,1.0\}$, totaling 16 configurations. The pair selected on the default task of each environment is reused on its other four tasks. For each D4RL dataset, we instead evaluate
$\alpha_{\mathrm{NF}}\in\{10^{-3},3\times10^{-3},10^{-2},3\times10^{-2}\}$ and
$\alpha_{\mathrm{aux}}\in\{0.005,0.01,0.03,0.1\}$, again totaling 16 configurations. Thus, D4RL changes only the candidate coefficient values and selects a pair per dataset; the shared recipe remains unchanged. Table~\ref{tab:bc-coefficients} reports these selected pairs; the additional OGBench variants and their extended grids are described with Table~\ref{tab:extended-ogbench-hyperparameters}.

Intermediate diagnostic evaluations logged during training are not used to select configurations. These interactions only choose hyperparameters; policy optimization never adds transitions to the offline datasets. The procedure is therefore offline RL with environment-assisted benchmark model selection, not fully offline model selection.

\begin{table}[!htbp]
\caption{Selected behavior-regularization coefficients. OGBench coefficients are selected on each category's default task and transferred to its other four tasks; D4RL coefficients are selected per dataset.}
\label{tab:bc-coefficients}
\centering
\footnotesize
\setlength{\tabcolsep}{6pt}
\begin{adjustbox}{max width=\linewidth}
\begin{tabular}{@{}llcc@{}}
\toprule
\textbf{Suite} & \textbf{Category / dataset} & $\boldsymbol{\alpha_{\mathrm{NF}}}$ & $\boldsymbol{\alpha_{\mathrm{aux}}}$ \\
\midrule
OGBench & AntMaze Large & $10^{-4}$ & $0.1$ \\
OGBench & AntMaze Giant & $10^{-4}$ & $0.03$ \\
OGBench & HumanoidMaze Medium & $10^{-4}$ & $0.03$ \\
OGBench & HumanoidMaze Large & $10^{-4}$ & $0.03$ \\
OGBench & AntSoccer Arena & $3\!\times\!10^{-3}$ & $0.1$ \\
OGBench & Cube Single & $10^{-4}$ & $1.0$ \\
OGBench & Cube Double & $10^{-3}$ & $0.1$ \\
OGBench & Scene & $10^{-3}$ & $0.5$ \\
OGBench & Puzzle $3\times3$ & $10^{-2}$ & $0.5$ \\
OGBench & Puzzle $4\times4$ & $10^{-4}$ & $0.1$ \\
\midrule
D4RL AntMaze & Umaze & $3\!\times\!10^{-2}$ & $0.1$ \\
D4RL AntMaze & Umaze Diverse & $10^{-3}$ & $0.1$ \\
D4RL AntMaze & Medium Play & $3\!\times\!10^{-3}$ & $0.01$ \\
D4RL AntMaze & Medium Diverse & $10^{-3}$ & $0.005$ \\
D4RL AntMaze & Large Play & $3\!\times\!10^{-3}$ & $0.005$ \\
D4RL AntMaze & Large Diverse & $10^{-3}$ & $0.005$ \\
\midrule
D4RL Adroit & Pen Human & $10^{-3}$ & $0.1$ \\
D4RL Adroit & Pen Cloned & $3\!\times\!10^{-2}$ & $0.1$ \\
D4RL Adroit & Door Human & $3\!\times\!10^{-3}$ & $0.01$ \\
D4RL Adroit & Door Cloned & $10^{-2}$ & $0.1$ \\
D4RL Adroit & Hammer Human & $10^{-2}$ & $0.01$ \\
D4RL Adroit & Hammer Cloned & $10^{-3}$ & $0.01$ \\
D4RL Adroit & Relocate Human & $10^{-2}$ & $0.01$ \\
D4RL Adroit & Relocate Cloned & $3\!\times\!10^{-2}$ & $0.1$ \\
\bottomrule
\end{tabular}
\end{adjustbox}

\end{table}

\begin{table}[!htbp]
\caption{Adaptation interfaces reconstructed from released configurations for the tasks compared in this paper. ReBRAC-v2 rows show the actual candidate grids evaluated; baseline rows show the Cartesian envelope of distinct values appearing in released final configurations, not the baselines' actual numbers of tuning trials. L and G denote Laplace and Gaussian kernels. Adroit includes the human and cloned datasets used in our comparison.}
\label{tab:released-tuning-envelopes}
\centering
\footnotesize
\setlength{\tabcolsep}{2pt}
\begin{adjustbox}{max width=\linewidth}
\begin{tabular}{@{}p{0.12\linewidth}p{0.11\linewidth}p{0.32\linewidth}p{0.10\linewidth}p{0.27\linewidth}@{}}
\toprule
\textbf{Method} & \textbf{Suite} & \textbf{Released values} & \textbf{Envelope} & \textbf{Other settings} \\
\midrule
ReBRAC-v2
& OGBench & $\alpha_{\mathrm{NF}}\!:\{10^{-4},\allowbreak 10^{-3},\allowbreak 3\times10^{-3},\allowbreak 10^{-2}\}$; \newline $\alpha_{\mathrm{aux}}\!:\{.03,\allowbreak .1,\allowbreak .5,\allowbreak 1\}$ & $4\times4=16$ & None \\
& D4RL & $\alpha_{\mathrm{NF}}\!:\{.001,\allowbreak .003,\allowbreak .01,\allowbreak .03\}$; \newline $\alpha_{\mathrm{aux}}\!:\{.005,\allowbreak .01,\allowbreak .03,\allowbreak .1\}$ & $4\times4=16$ & None \\
\midrule
GFP
& OGBench & $\alpha\!:\{.1,\allowbreak .3,\allowbreak 1,\allowbreak 3,\allowbreak 10\}$; \newline $\eta\!:\{10^{-5},\allowbreak 10^{-4},\allowbreak 10^{-3},\allowbreak 10^{-2},\allowbreak 10^{-1}\}$ & $5\times5=25$ & Batch size, discount, flow steps, $Q$/target/guidance aggregation; 1M-update horizon \\
& AntMaze & $\alpha\!:\{.03,\allowbreak .1\}$; \newline $\eta\!:\{10^{-5},\allowbreak 10^{-3}\}$ & $2\times2=4$ & 500K-update horizon \\
& Adroit & $\alpha\!:\{3,\allowbreak 10\}$; \newline $\eta\!:\{10^{-5},\allowbreak 10^{-4},\allowbreak 10^{-2}\}$ & $2\times3=6$ & 500K-update horizon \\
\midrule
FAC
& OGBench & $\alpha\!:\{.5,\allowbreak 1,\allowbreak 5\}$; \newline $\lambda\!:\{.1,\allowbreak .3,\allowbreak 1,\allowbreak 10\}$ & $3\times4=12$ & $Q$ aggregation, density threshold, likelihood estimator \\
& AntMaze & $\alpha\!:\{.5,\allowbreak 1,\allowbreak 5\}$; \newline $\lambda\!:\{.03,\allowbreak .1\}$ & $3\times2=6$ & Dataset-wide density threshold \\
& Adroit & $\alpha\!:\{.5,\allowbreak 1,\allowbreak 5,\allowbreak 10\}$; \newline $\lambda\!:\{.03,\allowbreak .1,\allowbreak .3,\allowbreak 1,\allowbreak 3,\allowbreak 10\}$ & $4\times6=24$ & Hutchinson likelihood estimator \\
\midrule
DriftQL
& OGBench & $T\!:\{.02,\allowbreak .2,\allowbreak .5,\allowbreak .8\}$; \newline $\alpha\!:\{10,\allowbreak 32,\allowbreak 50,\allowbreak 60,\allowbreak 65,\allowbreak 100,\allowbreak 250,\allowbreak 300\}$; kernel: \{L,G\} & $4\times8\times2=64$ & Discount, $Q$ aggregation, generation count, noise dimension; 1M-update horizon \\
& AntMaze & $T\!:\{.5\}$; \newline $\alpha\!:\{3,\allowbreak 5,\allowbreak 8,\allowbreak 12,\allowbreak 15\}$; kernel: \{L,G\} & $1\times5\times2=10$ & Discount, $Q$ aggregation; 500K-update horizon \\
& Adroit & $T\!:\{.05,\allowbreak .2\}$; \newline $\alpha\!:\{1500,\allowbreak 2000,\allowbreak 2500,\allowbreak 4500,\allowbreak 5000\}$; kernel: \{L,G\} & $2\times5\times2=20$ & Discount, $Q$ aggregation; 500K-update horizon \\
\bottomrule
\end{tabular}
\end{adjustbox}

\end{table}

\FloatBarrier
\begingroup
\footnotesize
\setlength{\tabcolsep}{1.7pt}
\renewcommand{\arraystretch}{1.08}
\begin{longtable}{@{}lrrrrrrrrrr@{}}
\caption{Full OGBench task-level results. We omit the common \texttt{-singletask-} and \texttt{-v0} portions of task identifiers; $^\ast$ marks the default task in each environment. ReBRAC-v2, FAC, GFP, and DriftQL entries are our final-checkpoint reruns. NF-RLBC results are reported by \citet{ghugare2025normalizing} and converted from the original 0--1 scale. The first average covers all 50 tasks and therefore omits NF-RLBC; the second recomputes every method on the same 30 tasks reported for NF-RLBC.}\label{tab:ogbench-full}\\
\toprule
\multicolumn{1}{c}{} & \multicolumn{3}{c}{\textbf{Gaussian Policies}} & \multicolumn{4}{c}{\textbf{Flow Policies}} & \textbf{Drift} & \multicolumn{2}{c}{\textbf{Normalizing Flows}} \\
\cmidrule(lr){2-4} \cmidrule(lr){5-8} \cmidrule(lr){9-9} \cmidrule(lr){10-11}
\textbf{Environment / Task} & \textbf{BC} & \textbf{IQL} & \textbf{ReBRAC} & \textbf{IFQL} & \textbf{FQL} & \textbf{FAC} & \textbf{GFP} & \textbf{DriftQL} & \textbf{NF-RLBC} & \textbf{ReBRAC-v2} \\
\midrule
\endfirsthead
\multicolumn{11}{l}{\tablename\ \thetable{} (continued)}\\
\toprule
\multicolumn{1}{c}{} & \multicolumn{3}{c}{\textbf{Gaussian Policies}} & \multicolumn{4}{c}{\textbf{Flow Policies}} & \textbf{Drift} & \multicolumn{2}{c}{\textbf{Normalizing Flows}} \\
\cmidrule(lr){2-4} \cmidrule(lr){5-8} \cmidrule(lr){9-9} \cmidrule(lr){10-11}
\textbf{Environment / Task} & \textbf{BC} & \textbf{IQL} & \textbf{ReBRAC} & \textbf{IFQL} & \textbf{FQL} & \textbf{FAC} & \textbf{GFP} & \textbf{DriftQL} & \textbf{NF-RLBC} & \textbf{ReBRAC-v2} \\
\midrule
\endhead
\midrule
\multicolumn{11}{r}{Continued on next page}\\
\endfoot
\bottomrule
\endlastfoot
\ogtask{AntMaze Large}{1$^\ast$} & 0 & 48 & 91 & 24.0 & 80.0 & 64.0 & \underline{93.4} & 91.6 & 76.4 & \textbf{98.8} \\
\ogtask{AntMaze Large}{2} & 6 & 42 & 88 & 8.0 & 57.0 & 51.6 & \underline{90.1} & 86.8 & 72.4 & \textbf{96.8} \\
\ogtask{AntMaze Large}{3} & 29 & 72 & 51 & 52.0 & 93.0 & 90.0 & 94.5 & \underline{95.6} & 92.4 & \textbf{100.0} \\
\ogtask{AntMaze Large}{4} & 8 & 51 & 84 & 18.0 & 80.0 & 2.0 & 91.8 & \underline{93.6} & 61.6 & \textbf{96.0} \\
\ogtask{AntMaze Large}{5} & 10 & 54 & 90 & 38.0 & 83.0 & 78.4 & \underline{95.3} & 74.8 & 60.8 & \textbf{97.6} \\
\midrule
\ogtask{AntMaze Giant}{1$^\ast$} & 0 & 0 & 27 & 0.0 & 4.0 & 0.0 & \underline{30.0} & 22.0 & -- & \textbf{90.4} \\
\ogtask{AntMaze Giant}{2} & 0 & 1 & 16 & 0.0 & 9.0 & 0.8 & \underline{66.5} & \textbf{74.8} & -- & \textbf{74.8} \\
\ogtask{AntMaze Giant}{3} & 0 & 0 & \underline{34} & 0.0 & 0.0 & 0.0 & 14.8 & \textbf{57.2} & -- & 30.8 \\
\ogtask{AntMaze Giant}{4} & 0 & 0 & 5 & 0.0 & 14.0 & 0.0 & 0.0 & \underline{74.8} & -- & \textbf{94.8} \\
\ogtask{AntMaze Giant}{5} & 1 & 19 & 49 & 13.0 & 16.0 & 0.0 & 48.5 & \underline{77.2} & -- & \textbf{84.0} \\
\midrule
\ogtask{HumanoidMaze Medium}{1$^\ast$} & 1 & 32 & 16 & 69.0 & 19.0 & 56.4 & \underline{86.2} & 22.4 & 3.6 & \textbf{98.0} \\
\ogtask{HumanoidMaze Medium}{2} & 1 & 41 & 18 & 85.0 & \underline{94.0} & 87.2 & 79.2 & 80.8 & 18.8 & \textbf{97.2} \\
\ogtask{HumanoidMaze Medium}{3} & 6 & 25 & 36 & 49.0 & 74.0 & 79.6 & \underline{88.3} & 58.4 & 14.8 & \textbf{96.4} \\
\ogtask{HumanoidMaze Medium}{4} & 0 & 0 & 15 & 1.0 & 3.0 & 15.2 & 0.0 & \textbf{48.8} & 1.6 & \underline{44.0} \\
\ogtask{HumanoidMaze Medium}{5} & 2 & 66 & 24 & 98.0 & 97.0 & 98.0 & \underline{98.6} & 96.8 & 20.8 & \textbf{99.2} \\
\midrule
\ogtask{HumanoidMaze Large}{1$^\ast$} & 0 & 3 & 2 & 6.0 & 7.0 & 3.6 & 0.0 & \underline{8.4} & -- & \textbf{95.2} \\
\ogtask{HumanoidMaze Large}{2} & 0 & 0 & 0 & 0.0 & 0.0 & 0.0 & 0.0 & 0.0 & -- & \textbf{83.6} \\
\ogtask{HumanoidMaze Large}{3} & 1 & 7 & 8 & \textbf{48.0} & 11.0 & 12.0 & \underline{33.9} & 16.0 & -- & 26.8 \\
\ogtask{HumanoidMaze Large}{4} & 1 & 1 & 1 & 1.0 & \underline{2.0} & 0.0 & 0.8 & 0.4 & -- & \textbf{66.4} \\
\ogtask{HumanoidMaze Large}{5} & 0 & 1 & 2 & 0.0 & 1.0 & 4.4 & \underline{39.9} & 0.4 & -- & \textbf{88.8} \\
\midrule
\ogtask{AntSoccer Arena}{1} & 2 & 14 & 0 & 61.0 & 77.0 & 83.2 & 83.3 & \underline{84.8} & 60.0 & \textbf{94.8} \\
\ogtask{AntSoccer Arena}{2} & 2 & 17 & 0 & 75.0 & 88.0 & \textbf{95.6} & \underline{89.9} & 85.6 & 55.6 & 83.6 \\
\ogtask{AntSoccer Arena}{3} & 0 & 6 & 0 & 14.0 & \textbf{61.0} & 54.0 & \underline{59.7} & 50.8 & 42.0 & 57.6 \\
\ogtask{AntSoccer Arena}{4$^\ast$} & 1 & 3 & 0 & 16.0 & 39.0 & 46.0 & 45.1 & 47.2 & \underline{48.0} & \textbf{51.6} \\
\ogtask{AntSoccer Arena}{5} & 0 & 2 & 0 & 0.0 & 36.0 & \underline{48.8} & 46.1 & 42.4 & 33.6 & \textbf{49.2} \\
\midrule
\ogtask{Cube Single}{1} & 10 & 88 & 89 & 79.0 & \underline{97.0} & 92.8 & \textbf{98.1} & 92.4 & 44.0 & 93.2 \\
\ogtask{Cube Single}{2$^\ast$} & 3 & 85 & 92 & 73.0 & 97.0 & \underline{99.2} & \textbf{99.4} & 89.2 & 64.0 & 95.6 \\
\ogtask{Cube Single}{3} & 9 & 91 & 93 & 88.0 & 98.0 & \underline{98.8} & \textbf{100.0} & 95.2 & 76.8 & 95.6 \\
\ogtask{Cube Single}{4} & 2 & 73 & 92 & 79.0 & 94.0 & \underline{98.0} & \textbf{99.4} & 82.8 & 60.0 & 91.6 \\
\ogtask{Cube Single}{5} & 3 & 78 & 87 & 77.0 & \underline{93.0} & 88.0 & \textbf{95.6} & 78.8 & 61.6 & 90.8 \\
\midrule
\ogtask{Cube Double}{1} & 8 & 27 & 45 & 35.0 & \underline{61.0} & 53.2 & \textbf{74.1} & 46.8 & -- & 6.4 \\
\ogtask{Cube Double}{2$^\ast$} & 0 & 1 & 7 & 9.0 & \underline{36.0} & 20.0 & \textbf{37.1} & 14.4 & -- & 16.0 \\
\ogtask{Cube Double}{3} & 0 & 0 & 4 & 8.0 & \underline{22.0} & 20.8 & \textbf{37.3} & 8.4 & -- & 7.2 \\
\ogtask{Cube Double}{4} & 0 & 0 & 1 & 1.0 & \textbf{5.0} & 1.6 & \underline{2.8} & 2.0 & -- & 2.0 \\
\ogtask{Cube Double}{5} & 0 & 4 & 4 & 17.0 & 19.0 & \underline{34.4} & \textbf{49.3} & 17.2 & -- & 10.0 \\
\midrule
\ogtask{Scene}{1} & 19 & 94 & 95 & 98.0 & \textbf{100.0} & \textbf{100.0} & \textbf{100.0} & 99.2 & \textbf{100.0} & \underline{99.6} \\
\ogtask{Scene}{2$^\ast$} & 1 & 12 & 50 & 0.0 & 76.0 & \textbf{100.0} & \underline{93.4} & 88.8 & 88.0 & \textbf{100.0} \\
\ogtask{Scene}{3} & 1 & 32 & 55 & 54.0 & \underline{98.0} & 94.8 & 74.9 & 95.6 & \textbf{98.8} & 95.2 \\
\ogtask{Scene}{4} & 2 & 0 & 3 & 0.0 & 5.0 & 9.6 & 0.0 & 68.4 & \underline{85.6} & \textbf{95.6} \\
\ogtask{Scene}{5} & 0 & 0 & 0 & 0.0 & 0.0 & 0.0 & 0.0 & \underline{8.0} & 0.0 & \textbf{86.8} \\
\midrule
\ogtask{Puzzle 3x3}{1} & 5 & 33 & 97 & 94.0 & 90.0 & \textbf{100.0} & 96.5 & 97.2 & \underline{99.6} & \textbf{100.0} \\
\ogtask{Puzzle 3x3}{2} & 1 & 4 & 1 & 1.0 & 16.0 & \underline{98.8} & 0.4 & 0.8 & \textbf{99.6} & \textbf{99.6} \\
\ogtask{Puzzle 3x3}{3} & 1 & 3 & 3 & 0.0 & 10.0 & \underline{98.4} & 0.8 & 3.2 & 97.6 & \textbf{98.8} \\
\ogtask{Puzzle 3x3}{4$^\ast$} & 1 & 2 & 2 & 0.0 & 16.0 & 96.8 & 9.4 & 54.0 & \underline{99.6} & \textbf{100.0} \\
\ogtask{Puzzle 3x3}{5} & 1 & 3 & 5 & 0.0 & 16.0 & \textbf{100.0} & 9.2 & 46.0 & \underline{99.6} & \underline{99.6} \\
\midrule
\ogtask{Puzzle 4x4}{1} & 1 & 12 & 26 & 49.0 & 34.0 & 50.8 & 60.8 & \underline{69.2} & -- & \textbf{90.4} \\
\ogtask{Puzzle 4x4}{2} & 0 & 7 & 12 & 4.0 & \underline{16.0} & 5.2 & 15.5 & 3.6 & -- & \textbf{37.6} \\
\ogtask{Puzzle 4x4}{3} & 0 & 9 & 15 & 50.0 & 18.0 & \underline{52.8} & 45.5 & 48.8 & -- & \textbf{78.0} \\
\ogtask{Puzzle 4x4}{4$^\ast$} & 0 & 5 & 10 & 21.0 & 11.0 & \underline{21.2} & 19.6 & 10.0 & -- & \textbf{30.0} \\
\ogtask{Puzzle 4x4}{5} & 0 & 4 & 7 & 2.0 & 7.0 & 2.4 & \underline{10.8} & 3.6 & -- & \textbf{25.2} \\
\midrule
\textbf{\texttt{Average (50 tasks)}} & 2.8 & 23.4 & 31.0 & 30.3 & 43.6 & 50.2 & 52.1 & \underline{52.3} & -- & \textbf{74.8} \\
\textbf{\texttt{Average (NF-RLBC 30 tasks)}} & 4.2 & 35.7 & 42.6 & 41.7 & 62.9 & \underline{74.2} & 67.3 & 68.7 & 61.2 & \textbf{90.1} \\
\end{longtable}
\endgroup

\FloatBarrier
\section{Extended OGBench Evaluation}
\label{sec:extended-ogbench}

The strong results on the common task subset motivate a broader evaluation intended to make ReBRAC-v2 a useful baseline for future offline RL research. We additionally evaluate 185 tasks from 37 state-based OGBench dataset variants beyond the 50-task main comparison. Together, these cover all 235 state-based single-task problems across the 47 locomotion and manipulation dataset variants in the original OGBench suite \citep{park2024ogbench}. Coverage comprises 8 PointMaze, 11 AntMaze, 6 HumanoidMaze, 4 AntSoccer, 8 Cube, 8 Puzzle, and 2 Scene dataset variants, each with five tasks. It excludes visual observations, Powderworld, and later large-scale dataset extensions.

None of the compared baselines reports results across this full set. We therefore present the additional-task results separately rather than include them in the main comparative averages or RLiable plots. Table~\ref{tab:extended-ogbench} reports final-checkpoint scores; Table~\ref{tab:extended-ogbench-hyperparameters} records the corresponding cloning coefficients. These results and configurations provide concrete reference points for broader evaluations without requiring subsequent work to restrict itself to the common 50-task subset.

The additional-task average is 40.6. Performance is strong on AntMaze Medium, PointMaze Medium Navigate and Giant Navigate, Cube Single Noisy, and Puzzle $3\times3$ Noisy. However, several stitching datasets and the larger Cube tasks remain poorly solved or unsolved. Without a complete baseline comparison, these results characterize the recipe's broader coverage and failure cases rather than establish comparative superiority.

\begingroup
\footnotesize
\setlength{\tabcolsep}{2pt}
\renewcommand{\arraystretch}{1.08}
\begin{longtable}{@{}lrrrrr|r@{}}
\caption{Additional state-based OGBench results. Each task reports mean $\pm$ population standard deviation over 4--5 seeds, evaluated for 50 episodes at the final checkpoint after 1M updates. Averages weight tasks equally. We omit the common \texttt{-singletask-} and \texttt{-v0} suffixes.}\label{tab:extended-ogbench}\\
\toprule
\textbf{Dataset variant} & \textbf{Task 1} & \textbf{Task 2} & \textbf{Task 3} & \textbf{Task 4} & \textbf{Task 5} & \textbf{Avg.} \\
\midrule
\endfirsthead
\multicolumn{7}{l}{\tablename\ \thetable{} (continued)}\\
\toprule
\textbf{Dataset variant} & \textbf{Task 1} & \textbf{Task 2} & \textbf{Task 3} & \textbf{Task 4} & \textbf{Task 5} & \textbf{Avg.} \\
\midrule
\endhead
\midrule
\multicolumn{7}{r}{Continued on next page}\\
\endfoot
\bottomrule
\endlastfoot
antmaze-giant-stitch & $0.4\pm0.8$ & $0.0\pm0.0$ & $0.0\pm0.0$ & $0.0\pm0.0$ & $9.6\pm4.5$ & 2.0 \\
antmaze-large-explore & $100.0\pm0.0$ & $20.0\pm9.5$ & $100.0\pm0.0$ & $99.6\pm0.8$ & $100.0\pm0.0$ & 83.9 \\
antmaze-large-stitch & $68.4\pm25.5$ & $4.0\pm2.2$ & $9.6\pm9.3$ & $0.0\pm0.0$ & $45.6\pm5.4$ & 25.5 \\
antmaze-medium-explore & $100.0\pm0.0$ & $100.0\pm0.0$ & $100.0\pm0.0$ & $100.0\pm0.0$ & $100.0\pm0.0$ & 100.0 \\
antmaze-medium-navigate & $99.2\pm1.0$ & $98.4\pm1.5$ & $98.0\pm1.3$ & $97.2\pm2.0$ & $98.4\pm1.5$ & 98.2 \\
antmaze-medium-stitch & $99.6\pm0.8$ & $100.0\pm0.0$ & $85.2\pm7.7$ & $90.4\pm4.1$ & $100.0\pm0.0$ & 95.0 \\
antmaze-teleport-explore & $46.8\pm8.4$ & $99.6\pm0.8$ & $52.8\pm9.9$ & $50.4\pm2.7$ & $35.2\pm18.2$ & 57.0 \\
antmaze-teleport-navigate & $40.8\pm7.2$ & $45.2\pm4.1$ & $46.8\pm7.3$ & $50.0\pm6.1$ & $47.6\pm2.7$ & 46.1 \\
antmaze-teleport-stitch & $47.2\pm8.5$ & $45.2\pm7.9$ & $42.8\pm5.7$ & $53.6\pm9.9$ & $52.0\pm4.2$ & 48.2 \\
\midrule
antsoccer-arena-stitch & $0.8\pm1.0$ & $0.4\pm0.8$ & $0.0\pm0.0$ & $0.0\pm0.0$ & $0.0\pm0.0$ & 0.2 \\
antsoccer-medium-navigate & $37.2\pm9.4$ & $1.2\pm2.4$ & $0.0\pm0.0$ & $14.8\pm7.2$ & $35.6\pm9.8$ & 17.8 \\
antsoccer-medium-stitch & $0.0\pm0.0$ & $0.0\pm0.0$ & $0.0\pm0.0$ & $0.0\pm0.0$ & $0.0\pm0.0$ & 0.0 \\
\midrule
cube-double-noisy & $29.2\pm7.0$ & $6.4\pm5.0$ & $6.8\pm4.1$ & $4.8\pm1.6$ & $2.8\pm2.0$ & 10.0 \\
cube-quadruple-noisy & $0.0\pm0.0$ & $0.0\pm0.0$ & $0.0\pm0.0$ & $0.0\pm0.0$ & $0.0\pm0.0$ & 0.0 \\
cube-quadruple-play & $0.0\pm0.0$ & $0.0\pm0.0$ & $0.0\pm0.0$ & $0.0\pm0.0$ & $0.0\pm0.0$ & 0.0 \\
cube-single-noisy & $99.6\pm0.8$ & $100.0\pm0.0$ & $100.0\pm0.0$ & $100.0\pm0.0$ & $100.0\pm0.0$ & 99.9 \\
cube-triple-noisy & $9.2\pm6.4$ & $0.0\pm0.0$ & $0.0\pm0.0$ & $0.0\pm0.0$ & $0.0\pm0.0$ & 1.8 \\
cube-triple-play & $2.0\pm2.2$ & $0.0\pm0.0$ & $0.0\pm0.0$ & $0.0\pm0.0$ & $0.0\pm0.0$ & 0.4 \\
\midrule
humanoidmaze-giant-navigate & $81.2\pm4.1$ & $0.0\pm0.0$ & $20.0\pm34.1$ & $74.0\pm37.1$ & $17.6\pm35.2$ & 38.6 \\
humanoidmaze-giant-stitch & $0.0\pm0.0$ & $2.8\pm2.4$ & $0.0\pm0.0$ & $4.0\pm3.3$ & $44.0\pm10.7$ & 10.2 \\
humanoidmaze-large-stitch & $5.6\pm4.1$ & $0.0\pm0.0$ & $49.6\pm34.1$ & $24.0\pm1.8$ & $9.2\pm4.8$ & 17.7 \\
humanoidmaze-medium-stitch & $83.6\pm4.8$ & $94.0\pm2.8$ & $98.0\pm3.1$ & $82.4\pm17.5$ & $98.8\pm1.6$ & 91.4 \\
\midrule
pointmaze-giant-navigate & $100.0\pm0.0$ & $100.0\pm0.0$ & $100.0\pm0.0$ & $97.2\pm3.5$ & $98.8\pm1.6$ & 99.2 \\
pointmaze-giant-stitch & $0.0\pm0.0$ & $0.0\pm0.0$ & $0.0\pm0.0$ & $0.0\pm0.0$ & $0.0\pm0.0$ & 0.0 \\
pointmaze-large-navigate & $100.0\pm0.0$ & $22.4\pm27.5$ & $100.0\pm0.0$ & $100.0\pm0.0$ & $100.0\pm0.0$ & 84.5 \\
pointmaze-large-stitch & $0.0\pm0.0$ & $0.0\pm0.0$ & $0.0\pm0.0$ & $0.0\pm0.0$ & $0.0\pm0.0$ & 0.0 \\
pointmaze-medium-navigate & $100.0\pm0.0$ & $100.0\pm0.0$ & $100.0\pm0.0$ & $100.0\pm0.0$ & $100.0\pm0.0$ & 100.0 \\
pointmaze-medium-stitch & $80.0\pm40.0$ & $71.2\pm39.5$ & $100.0\pm0.0$ & $99.6\pm0.8$ & $100.0\pm0.0$ & 90.2 \\
pointmaze-teleport-navigate & $7.6\pm15.2$ & $20.8\pm25.8$ & $30.8\pm25.2$ & $10.8\pm16.8$ & $23.6\pm8.9$ & 18.7 \\
pointmaze-teleport-stitch & $0.0\pm0.0$ & $0.0\pm0.0$ & $34.4\pm19.9$ & $13.2\pm19.1$ & $23.2\pm15.2$ & 14.2 \\
\midrule
puzzle-3x3-noisy & $100.0\pm0.0$ & $100.0\pm0.0$ & $100.0\pm0.0$ & $100.0\pm0.0$ & $100.0\pm0.0$ & 100.0 \\
puzzle-4x4-noisy & $68.0\pm14.0$ & $0.0\pm0.0$ & $20.0\pm14.3$ & $24.0\pm13.1$ & $0.0\pm0.0$ & 22.4 \\
puzzle-4x5-noisy & $96.8\pm3.7$ & $0.0\pm0.0$ & $0.0\pm0.0$ & $0.0\pm0.0$ & $0.0\pm0.0$ & 19.4 \\
puzzle-4x5-play & $100.0\pm0.0$ & $0.0\pm0.0$ & $0.0\pm0.0$ & $0.0\pm0.0$ & $0.0\pm0.0$ & 20.0 \\
puzzle-4x6-noisy & $0.8\pm1.6$ & $0.0\pm0.0$ & $0.0\pm0.0$ & $0.0\pm0.0$ & $0.0\pm0.0$ & 0.2 \\
puzzle-4x6-play & $25.2\pm5.2$ & $2.8\pm2.0$ & $0.4\pm0.8$ & $0.0\pm0.0$ & $0.0\pm0.0$ & 5.7 \\
\midrule
scene-noisy & $100.0\pm0.0$ & $89.2\pm7.1$ & $89.6\pm8.2$ & $72.4\pm14.2$ & $67.2\pm10.6$ & 83.7 \\
\midrule
\multicolumn{6}{l|}{\textbf{Average over 185 additional tasks}} & 40.6 \\
\end{longtable}
\endgroup

\FloatBarrier
\subsection{Hyperparameters for Additional Dataset Variants}

We first apply the same 16-point behavior-regularization grid used for the common OGBench tasks, then extend the coefficient grid for some additional datasets. Among the selected configurations, likelihood coefficients $10^{-5}$ and $3\times10^{-5}$ and auxiliary coefficients $0.003$ and $0.01$ lie outside the original OGBench grid. These are observed selected values, not a reconstruction of every candidate tested. Consequently, the 16-configuration adaptation budget stated for the main comparison does not describe the total search budget for every additional dataset. Within each additional dataset variant, the selected pair is reused across its five tasks. Architecture, optimizer, regularization, warm-ups, discount, and inference settings retain the original shared recipe; evaluation uses 1M updates, 50 episodes, and training seeds from the range 10--14.

\begingroup
\small
\setlength{\tabcolsep}{2pt}
\renewcommand{\arraystretch}{1.08}
\begin{longtable}{@{}lcc@{}}
\caption{Recorded behavior-regularization coefficients for the additional OGBench dataset variants in Table~\ref{tab:extended-ogbench}. Each pair is shared across all five tasks and seeds of its dataset variant. All other recorded training and inference settings are shared and match the existing ReBRAC-v2 recipe (Table~\ref{tab:shared-hyperparameters}).}\label{tab:extended-ogbench-hyperparameters}\\
\toprule
\textbf{Dataset variant} & $\boldsymbol{\alpha_{\mathrm{NF}}}$ & $\boldsymbol{\alpha_{\mathrm{aux}}}$ \\
\midrule
\endfirsthead
\multicolumn{3}{l}{\tablename\ \thetable{} (continued)}\\
\toprule
\textbf{Dataset variant} & $\boldsymbol{\alpha_{\mathrm{NF}}}$ & $\boldsymbol{\alpha_{\mathrm{aux}}}$ \\
\midrule
\endhead
\midrule
\multicolumn{3}{r}{Continued on next page}\\
\endfoot
\bottomrule
\endlastfoot
antmaze-giant-stitch & 1e-05 & 0.003 \\
antmaze-large-explore & 0.0001 & 0.03 \\
antmaze-large-stitch & 0.0001 & 0.003 \\
antmaze-medium-explore & 0.0001 & 0.03 \\
antmaze-medium-navigate & 0.0001 & 0.5 \\
antmaze-medium-stitch & 0.0001 & 0.03 \\
antmaze-teleport-explore & 3e-05 & 0.003 \\
antmaze-teleport-navigate & 1e-05 & 0.01 \\
antmaze-teleport-stitch & 1e-05 & 0.03 \\
\midrule
antsoccer-arena-stitch & 0.001 & 0.5 \\
antsoccer-medium-navigate & 1e-05 & 0.03 \\
antsoccer-medium-stitch & 0.01 & 1 \\
\midrule
cube-double-noisy & 0.003 & 0.03 \\
cube-quadruple-noisy & 0.01 & 0.5 \\
cube-quadruple-play & 0.001 & 0.5 \\
cube-single-noisy & 0.003 & 0.1 \\
cube-triple-noisy & 0.01 & 0.5 \\
cube-triple-play & 0.001 & 0.5 \\
\midrule
humanoidmaze-giant-navigate & 3e-05 & 0.03 \\
humanoidmaze-giant-stitch & 1e-05 & 0.03 \\
humanoidmaze-large-stitch & 0.0001 & 0.03 \\
humanoidmaze-medium-stitch & 0.0001 & 0.03 \\
\midrule
pointmaze-giant-navigate & 0.003 & 0.01 \\
pointmaze-giant-stitch & 0.01 & 0.03 \\
pointmaze-large-navigate & 0.01 & 0.03 \\
pointmaze-large-stitch & 0.01 & 0.03 \\
pointmaze-medium-navigate & 0.01 & 0.03 \\
pointmaze-medium-stitch & 0.001 & 0.01 \\
pointmaze-teleport-navigate & 0.001 & 0.03 \\
pointmaze-teleport-stitch & 0.01 & 0.03 \\
\midrule
puzzle-3x3-noisy & 0.003 & 0.1 \\
puzzle-4x4-noisy & 0.01 & 0.003 \\
puzzle-4x5-noisy & 0.003 & 0.01 \\
puzzle-4x5-play & 0.003 & 0.03 \\
puzzle-4x6-noisy & 0.003 & 0.01 \\
puzzle-4x6-play & 0.001 & 0.5 \\
\midrule
scene-noisy & 0.001 & 0.1 \\
\end{longtable}
\endgroup

\FloatBarrier
\section{Reproducibility Notes}

\paragraph{Random seeds.}
Bayesian development uses training seed 0, and behavior-regularization tuning uses seeds 0--3. Final OGBench evaluation uses seeds 10--14, while final D4RL evaluation uses seeds 10--19. Ablations use consecutive seeds beginning at 10, with the prefix length matching the reported number of runs (e.g., 10--13 for four runs). The released implementation initializes Python and JAX training randomness from the training seed and fixes the evaluation-environment seed to 42.

\paragraph{Compute and software.}
Experiments were distributed over heterogeneous academic compute acquired opportunistically, using NVIDIA H100 and NVIDIA TITAN RTX GPUs alongside different CPU models and memory capacities. ReBRAC-v2 uses one million gradient updates per run, and the public code repository provides a pinned Conda environment with the software and library versions used by the released implementation.

The separate inference-runtime benchmark uses an NVIDIA GeForce RTX 3090; its measurement protocol and software versions are detailed in the supplementary material subsection ``Fused Inference Runtime.''

\paragraph{Code release.}
The public repository linked in the abstract contains the ReBRAC-v2 training and evaluation implementations, final benchmark configurations, dataset-download helper, pinned environment, and result CSVs. The code is available under the Apache 2.0 license.

\FloatBarrier
\section{Baseline and Aggregate-Plot Protocols}
\label{sec:baseline-protocols}

\paragraph{NF-RLBC Coverage.}
We additionally include the NF-RLBC results reported by \citet{ghugare2025normalizing}. NF-RLBC instantiates a maximum-entropy RL+BC objective with a six-block normalizing-flow policy and is evaluated with five seeds. Its paper reports 30 OGBench tasks spanning AntMaze Large, AntSoccer Arena, Cube Single, HumanoidMaze Medium, Puzzle $3\times3$, and Scene; Supplementary Table~\ref{tab:ogbench-full} converts the reported means from the original 0--1 scale to our 0--100 convention. The table reports both the standard 50-task average and a fair comparison that recomputes every method on exactly these 30 tasks. On the matched subset, NF-RLBC averages 61.2, the next strongest baseline FAC averages 74.2, and ReBRAC-v2 averages 90.1; ReBRAC-v2 also obtains the higher category average in all six categories. Because NF-RLBC does not report the other 20 tasks and we do not have its seed-level scores, we exclude it from the 50-task main-table average and the RLiable plots.

\paragraph{Why Final-Checkpoint Evaluation?}
We use the policy produced at the end of a fixed training budget as the primary estimand. This rule is pre-specified, directly corresponds to the output of the training procedure, and introduces no checkpoint-selection window. It is also consistent with established pre-OGBench evaluation practice in TD3+BC, CQL, SAC-N/EDAC, and ReBRAC \citep{fujimoto2021minimalist,kumar2020conservative,an2021uncertainty,tarasov2024revisiting}. CORL explicitly reports the performance of the ``last trained policy'' in its fixed-budget D4RL benchmark, including its IQL and ReBRAC implementations \citep{kostrikov2021offline,tarasov2024corl}. Averaging several late checkpoints is not intrinsically invalid, but it evaluates a different quantity: performance over a chosen region of the learning trajectory. It therefore requires fixing the window, checkpoint count, spacing, and weighting. Such averaging can smooth transient variation, but it can also conceal late degradation or oscillation and can change method rankings relative to the final policy. Multi-seed evaluation addresses variability across independent runs; it does not characterize temporal instability within each run. We therefore rerun recent baselines under one final-checkpoint rule rather than directly comparing estimates with different meanings.

\paragraph{Source Protocols and Reproduction Differences.}
The GFP and DriftQL papers use eight seeds. On OGBench, they train for one million updates and average the checkpoints at 800K, 900K, and one million updates; on D4RL, they train for 500K updates and evaluate the final checkpoint. FAC reports eight seeds and 25 evaluation episodes, but its checkpoint-selection rule and complete evaluation protocol are not specified clearly enough to reconstruct. Each reproduction preserves the training horizon specified by the corresponding source paper or released configuration. In particular, baselines originally trained for 500K D4RL updates are rerun for 500K rather than one million updates. We align final-checkpoint selection, evaluation seeds, and episode counts without extending the original optimization budget.

For GFP, changing to our protocol produces similar aggregate results: the OGBench average changes from 51.8 to 52.1, D4RL AntMaze from 83.1 to 83.3, and D4RL Adroit from 20.3 to 21.4. DriftQL changes from 53.9 to 52.3 on OGBench, from 84.2 to 77.3 on AntMaze, and from 14.5 to 12.5 on Adroit. FAC changes more substantially, from 60.3 to 50.2 on OGBench, 90.5 to 85.3 on AntMaze, and 25.9 to 21.9 on Adroit. Substituting FAC's published values would not change its ordering relative to ReBRAC-v2 on OGBench (60.3 versus 74.8) or Adroit (25.9 versus 33.6), but would place FAC slightly higher on AntMaze (90.5 versus 90.2). We used the released implementations and hyperparameters, but the available evidence cannot distinguish among undocumented evaluation details, software or environment differences, stochastic variation, or mismatches between reported and released configurations. We therefore report discrepancies without assigning a cause.

\paragraph{ReBRAC-v2 under Late-Checkpoint Averaging.}
To isolate whether checkpoint selection explains the OGBench advantage, we also evaluate ReBRAC-v2 with the protocol used by GFP and DriftQL: within each seed, we average scores at 800K, 900K, and one million updates. Table~\ref{tab:rebrac-v2-checkpoint-average} reports the resulting task-level comparison. The independently tracked final-checkpoint average is 75.0, close to the 74.8 reported in the main results, while late-checkpoint averaging yields 73.1, a decrease of 2.0 points. The largest category-level change is on HumanoidMaze Large (74.3 to 62.0), followed by AntMaze Giant (77.9 to 73.3); each of the remaining eight categories changes by at most 2.1 points, and AntSoccer Arena and Cube Double improve slightly. Most importantly, under the same averaging rule, ReBRAC-v2 remains 19.2 points above the published DriftQL average (53.9) and 21.3 points above GFP (51.8). It also remains 12.8 points above FAC's published OGBench average (60.3), although FAC does not specify an equivalent checkpoint-selection protocol. Thus, the aggregate advantage does not arise from using the final checkpoint rather than late-checkpoint averaging.

\begingroup
\small
\setlength{\tabcolsep}{2pt}
\renewcommand{\arraystretch}{1.08}
\begin{longtable}{@{}lrrr@{}}
\caption{Effect of checkpoint selection on ReBRAC-v2 OGBench scores. Entries are mean $\pm$ population standard deviation across five seeds. The averaged protocol first averages the 800K, 900K, and 1M scores within each seed. The final-checkpoint and averaged estimates are then aggregated identically across tasks.}\label{tab:rebrac-v2-checkpoint-average}\\
\toprule
\textbf{Environment / Task} & \textbf{Final (1M)} & \textbf{Avg. (800K--1M)} & \textbf{$\Delta$} \\
\midrule
\endfirsthead
\multicolumn{4}{l}{\tablename\ \thetable{} (continued)}\\
\toprule
\textbf{Environment / Task} & \textbf{Final (1M)} & \textbf{Avg. (800K--1M)} & \textbf{$\Delta$} \\
\midrule
\endhead
\midrule
\multicolumn{4}{r}{Continued on next page}\\
\endfoot
\bottomrule
\endlastfoot
\ogtask{AntMaze Giant}{1} & $87.6 \pm 5.6$ & $88.4 \pm 1.9$ & +0.8 \\
\ogtask{AntMaze Giant}{2} & $92.8 \pm 3.2$ & $94.1 \pm 2.5$ & +1.3 \\
\ogtask{AntMaze Giant}{3} & $53.2 \pm 43.6$ & $37.5 \pm 35.0$ & -15.7 \\
\ogtask{AntMaze Giant}{4} & $75.2 \pm 37.8$ & $73.6 \pm 36.9$ & -1.6 \\
\ogtask{AntMaze Giant}{5} & $80.8 \pm 6.5$ & $72.9 \pm 11.0$ & -7.9 \\
\midrule
\ogtask{AntMaze Large}{1} & $99.2 \pm 1.0$ & $99.2 \pm 0.8$ & +0.0 \\
\ogtask{AntMaze Large}{2} & $96.8 \pm 3.7$ & $96.7 \pm 2.3$ & -0.1 \\
\ogtask{AntMaze Large}{3} & $99.2 \pm 1.0$ & $98.5 \pm 0.7$ & -0.7 \\
\ogtask{AntMaze Large}{4} & $98.4 \pm 1.5$ & $97.5 \pm 2.2$ & -0.9 \\
\ogtask{AntMaze Large}{5} & $98.8 \pm 1.0$ & $98.4 \pm 0.7$ & -0.4 \\
\midrule
\ogtask{AntSoccer Arena}{1} & $92.0 \pm 1.3$ & $92.3 \pm 1.8$ & +0.3 \\
\ogtask{AntSoccer Arena}{2} & $85.6 \pm 3.4$ & $90.7 \pm 2.1$ & +5.1 \\
\ogtask{AntSoccer Arena}{3} & $44.8 \pm 6.1$ & $47.9 \pm 6.1$ & +3.1 \\
\ogtask{AntSoccer Arena}{4} & $49.6 \pm 5.9$ & $50.0 \pm 3.5$ & +0.4 \\
\ogtask{AntSoccer Arena}{5} & $52.8 \pm 4.8$ & $48.8 \pm 3.6$ & -4.0 \\
\midrule
\ogtask{Cube Double}{1} & $9.6 \pm 3.2$ & $8.8 \pm 1.8$ & -0.8 \\
\ogtask{Cube Double}{2} & $8.8 \pm 3.7$ & $11.3 \pm 3.3$ & +2.5 \\
\ogtask{Cube Double}{3} & $14.8 \pm 4.7$ & $15.3 \pm 4.4$ & +0.5 \\
\ogtask{Cube Double}{4} & $1.6 \pm 1.5$ & $1.7 \pm 1.0$ & +0.1 \\
\ogtask{Cube Double}{5} & $11.2 \pm 8.3$ & $12.8 \pm 10.1$ & +1.6 \\
\midrule
\ogtask{Cube Single}{1} & $90.0 \pm 3.3$ & $91.5 \pm 3.1$ & +1.5 \\
\ogtask{Cube Single}{2} & $93.6 \pm 2.3$ & $92.8 \pm 3.4$ & -0.8 \\
\ogtask{Cube Single}{3} & $95.6 \pm 2.3$ & $95.7 \pm 1.5$ & +0.1 \\
\ogtask{Cube Single}{4} & $87.2 \pm 4.7$ & $87.7 \pm 4.4$ & +0.5 \\
\ogtask{Cube Single}{5} & $89.6 \pm 6.2$ & $88.1 \pm 4.0$ & -1.5 \\
\midrule
\ogtask{HumanoidMaze Large}{1} & $96.4 \pm 2.3$ & $96.0 \pm 1.3$ & -0.4 \\
\ogtask{HumanoidMaze Large}{2} & $79.2 \pm 7.5$ & $73.9 \pm 4.3$ & -5.3 \\
\ogtask{HumanoidMaze Large}{3} & $26.0 \pm 15.2$ & $25.2 \pm 13.4$ & -0.8 \\
\ogtask{HumanoidMaze Large}{4} & $86.0 \pm 5.9$ & $32.1 \pm 3.6$ & -53.9 \\
\ogtask{HumanoidMaze Large}{5} & $84.0 \pm 8.9$ & $82.7 \pm 7.1$ & -1.3 \\
\midrule
\ogtask{HumanoidMaze Medium}{1} & $97.6 \pm 1.5$ & $96.4 \pm 2.5$ & -1.2 \\
\ogtask{HumanoidMaze Medium}{2} & $96.0 \pm 2.8$ & $95.9 \pm 1.9$ & -0.1 \\
\ogtask{HumanoidMaze Medium}{3} & $96.8 \pm 2.7$ & $91.1 \pm 4.8$ & -5.7 \\
\ogtask{HumanoidMaze Medium}{4} & $37.6 \pm 13.2$ & $34.5 \pm 11.2$ & -3.1 \\
\ogtask{HumanoidMaze Medium}{5} & $100.0 \pm 0.0$ & $99.5 \pm 0.5$ & -0.5 \\
\midrule
\ogtask{Puzzle 3x3}{1} & $100.0 \pm 0.0$ & $99.9 \pm 0.3$ & -0.1 \\
\ogtask{Puzzle 3x3}{2} & $100.0 \pm 0.0$ & $99.9 \pm 0.3$ & -0.1 \\
\ogtask{Puzzle 3x3}{3} & $99.2 \pm 1.6$ & $99.2 \pm 0.8$ & +0.0 \\
\ogtask{Puzzle 3x3}{4} & $100.0 \pm 0.0$ & $99.6 \pm 0.5$ & -0.4 \\
\ogtask{Puzzle 3x3}{5} & $99.6 \pm 0.8$ & $99.7 \pm 0.5$ & +0.1 \\
\midrule
\ogtask{Puzzle 4x4}{1} & $87.6 \pm 9.4$ & $90.7 \pm 3.3$ & +3.1 \\
\ogtask{Puzzle 4x4}{2} & $38.8 \pm 13.5$ & $37.2 \pm 11.6$ & -1.6 \\
\ogtask{Puzzle 4x4}{3} & $83.2 \pm 11.4$ & $81.9 \pm 10.0$ & -1.3 \\
\ogtask{Puzzle 4x4}{4} & $38.0 \pm 17.0$ & $30.1 \pm 10.1$ & -7.9 \\
\ogtask{Puzzle 4x4}{5} & $20.0 \pm 13.2$ & $21.2 \pm 6.9$ & +1.2 \\
\midrule
\ogtask{Scene}{1} & $100.0 \pm 0.0$ & $99.7 \pm 0.5$ & -0.3 \\
\ogtask{Scene}{2} & $100.0 \pm 0.0$ & $99.9 \pm 0.3$ & -0.1 \\
\ogtask{Scene}{3} & $94.4 \pm 4.1$ & $93.6 \pm 2.2$ & -0.8 \\
\ogtask{Scene}{4} & $95.6 \pm 1.5$ & $92.8 \pm 1.1$ & -2.8 \\
\ogtask{Scene}{5} & $86.4 \pm 2.7$ & $87.7 \pm 2.0$ & +1.3 \\
\midrule
\textbf{Average (50 tasks)} & 75.0 & \textbf{73.1} & -2.0 \\
\end{longtable}
\endgroup

\begingroup
\small
\setlength{\tabcolsep}{2pt}
\renewcommand{\arraystretch}{1.08}
\begin{longtable}{@{}lrrrrrr@{}}
\caption{OGBench baseline reproduction table. Reported values are taken from the corresponding papers; ``Ours'' denotes evaluation under the ReBRAC-v2 final-checkpoint protocol. $^\ast$ marks the default task in each environment.}\label{tab:ogbench-reproductions}\\
\toprule
& \multicolumn{2}{c}{\textbf{FAC}} & \multicolumn{2}{c}{\textbf{GFP}} & \multicolumn{2}{c}{\textbf{DriftQL}} \\
\cmidrule(lr){2-3} \cmidrule(lr){4-5} \cmidrule(lr){6-7}
\textbf{Environment / Task} & \textbf{Reported} & \textbf{Ours} & \textbf{Reported} & \textbf{Ours} & \textbf{Reported} & \textbf{Ours} \\
\midrule
\endfirsthead
\multicolumn{7}{l}{\tablename\ \thetable{} (continued)}\\
\toprule
& \multicolumn{2}{c}{\textbf{FAC}} & \multicolumn{2}{c}{\textbf{GFP}} & \multicolumn{2}{c}{\textbf{DriftQL}} \\
\cmidrule(lr){2-3} \cmidrule(lr){4-5} \cmidrule(lr){6-7}
\textbf{Environment / Task} & \textbf{Reported} & \textbf{Ours} & \textbf{Reported} & \textbf{Ours} & \textbf{Reported} & \textbf{Ours} \\
\midrule
\endhead
\midrule
\multicolumn{7}{r}{Continued on next page}\\
\endfoot
\bottomrule
\endlastfoot
\ogtask{AntMaze Large}{1$^\ast$} & 94.0 & 64.0 & 95.4 & 93.4 & 95 & 91.6 \\
\ogtask{AntMaze Large}{2} & 86.0 & 51.6 & 92.2 & 90.1 & 85 & 86.8 \\
\ogtask{AntMaze Large}{3} & 97.5 & 90.0 & 95.6 & 94.5 & 97 & 95.6 \\
\ogtask{AntMaze Large}{4} & 89.5 & 2.0 & 90.6 & 91.8 & 91 & 93.6 \\
\ogtask{AntMaze Large}{5} & 96.0 & 78.4 & 95.0 & 95.3 & 92 & 74.8 \\
\midrule
\ogtask{AntMaze Giant}{1$^\ast$} & 6.5 & 0.0 & 12.6 & 30.0 & 32 & 22.0 \\
\ogtask{AntMaze Giant}{2} & 37.5 & 0.8 & 52.2 & 66.5 & 79 & 74.8 \\
\ogtask{AntMaze Giant}{3} & 0.5 & 0.0 & 13.7 & 14.8 & 43 & 57.2 \\
\ogtask{AntMaze Giant}{4} & 20.0 & 0.0 & 17.8 & 0.0 & 64 & 74.8 \\
\ogtask{AntMaze Giant}{5} & 50.5 & 0.0 & 43.2 & 48.5 & 85 & 77.2 \\
\midrule
\ogtask{HumanoidMaze Medium}{1$^\ast$} & 71.5 & 56.4 & 83.5 & 86.2 & 28 & 22.4 \\
\ogtask{HumanoidMaze Medium}{2} & 88.0 & 87.2 & 91.2 & 79.2 & 87 & 80.8 \\
\ogtask{HumanoidMaze Medium}{3} & 95.5 & 79.6 & 86.3 & 88.3 & 56 & 58.4 \\
\ogtask{HumanoidMaze Medium}{4} & 25.0 & 15.2 & 3.0 & 0.0 & 39 & 48.8 \\
\ogtask{HumanoidMaze Medium}{5} & 98.0 & 98.0 & 95.8 & 98.6 & 99 & 96.8 \\
\midrule
\ogtask{HumanoidMaze Large}{1$^\ast$} & 15.0 & 3.6 & 57.2 & 0.0 & 2 & 8.4 \\
\ogtask{HumanoidMaze Large}{2} & 0.0 & 0.0 & 0.1 & 0.0 & 0 & 0.0 \\
\ogtask{HumanoidMaze Large}{3} & 20.5 & 12.0 & 14.6 & 33.9 & 23 & 16.0 \\
\ogtask{HumanoidMaze Large}{4} & 4.5 & 0.0 & 3.7 & 0.8 & 1 & 0.4 \\
\ogtask{HumanoidMaze Large}{5} & 1.5 & 4.4 & 13.1 & 39.9 & 1 & 0.4 \\
\midrule
\ogtask{AntSoccer Arena}{1} & 82.0 & 83.2 & 77.0 & 83.3 & 79 & 84.8 \\
\ogtask{AntSoccer Arena}{2} & 93.5 & 95.6 & 91.2 & 89.9 & 91 & 85.6 \\
\ogtask{AntSoccer Arena}{3} & 62.5 & 54.0 & 51.9 & 59.7 & 60 & 50.8 \\
\ogtask{AntSoccer Arena}{4$^\ast$} & 53.0 & 46.0 & 40.2 & 45.1 & 48 & 47.2 \\
\ogtask{AntSoccer Arena}{5} & 47.5 & 48.8 & 29.1 & 46.1 & 48 & 42.4 \\
\midrule
\ogtask{Cube Single}{1} & 99.0 & 92.8 & 99.1 & 98.1 & 94 & 92.4 \\
\ogtask{Cube Single}{2$^\ast$} & 100.0 & 99.2 & 99.4 & 99.4 & 93 & 89.2 \\
\ogtask{Cube Single}{3} & 100.0 & 98.8 & 99.4 & 100.0 & 95 & 95.2 \\
\ogtask{Cube Single}{4} & 98.5 & 98.0 & 99.1 & 99.4 & 92 & 82.8 \\
\ogtask{Cube Single}{5} & 96.5 & 88.0 & 97.0 & 95.6 & 90 & 78.8 \\
\midrule
\ogtask{Cube Double}{1} & 60.0 & 53.2 & 76.1 & 74.1 & 49 & 46.8 \\
\ogtask{Cube Double}{2$^\ast$} & 37.5 & 20.0 & 53.3 & 37.1 & 23 & 14.4 \\
\ogtask{Cube Double}{3} & 31.5 & 20.8 & 43.3 & 37.3 & 9 & 8.4 \\
\ogtask{Cube Double}{4} & 4.0 & 1.6 & 7.1 & 2.8 & 3 & 2.0 \\
\ogtask{Cube Double}{5} & 32.5 & 34.4 & 56.3 & 49.3 & 43 & 17.2 \\
\midrule
\ogtask{Scene}{1} & 100.0 & 100.0 & 99.8 & 100.0 & 100 & 99.2 \\
\ogtask{Scene}{2$^\ast$} & 100.0 & 100.0 & 89.0 & 93.4 & 89 & 88.8 \\
\ogtask{Scene}{3} & 97.0 & 94.8 & 78.0 & 74.9 & 93 & 95.6 \\
\ogtask{Scene}{4} & 58.0 & 9.6 & 0.6 & 0.0 & 83 & 68.4 \\
\ogtask{Scene}{5} & 1.5 & 0.0 & 0.0 & 0.0 & 2 & 8.0 \\
\midrule
\ogtask{Puzzle 3x3}{1} & 100.0 & 100.0 & 94.8 & 96.5 & 87 & 97.2 \\
\ogtask{Puzzle 3x3}{2} & 100.0 & 98.8 & 0.3 & 0.4 & 39 & 0.8 \\
\ogtask{Puzzle 3x3}{3} & 100.0 & 98.4 & 0.9 & 0.8 & 20 & 3.2 \\
\ogtask{Puzzle 3x3}{4$^\ast$} & 100.0 & 96.8 & 5.4 & 9.4 & 10 & 54.0 \\
\ogtask{Puzzle 3x3}{5} & 100.0 & 100.0 & 14.1 & 9.2 & 19 & 46.0 \\
\midrule
\ogtask{Puzzle 4x4}{1} & 52.0 & 50.8 & 50.0 & 60.8 & 72 & 69.2 \\
\ogtask{Puzzle 4x4}{2} & 7.5 & 5.2 & 9.9 & 15.5 & 4 & 3.6 \\
\ogtask{Puzzle 4x4}{3} & 62.0 & 52.8 & 46.2 & 45.5 & 47 & 48.8 \\
\ogtask{Puzzle 4x4}{4$^\ast$} & 35.0 & 21.2 & 17.2 & 19.6 & 10 & 10.0 \\
\ogtask{Puzzle 4x4}{5} & 5.0 & 2.4 & 7.3 & 10.8 & 2 & 3.6 \\
\midrule
\textbf{\texttt{Average}} & 60.3 & 50.2 & 51.8 & 52.1 & 53.9 & 52.3 \\
\end{longtable}
\endgroup

\begin{table}[!htbp]
\caption{D4RL AntMaze baseline reproduction table. Reported values are taken from the corresponding papers; ``Ours'' denotes evaluation under the ReBRAC-v2 final-checkpoint protocol.}
\label{tab:d4rl-antmaze-reproductions}
\centering
\appendixtableformat
\setlength{\tabcolsep}{4pt}
\begin{adjustbox}{max width=\linewidth}
\begin{tabular}{@{}lrrrrrr@{}}
\toprule
& \multicolumn{2}{c}{\textbf{FAC}} & \multicolumn{2}{c}{\textbf{GFP}} & \multicolumn{2}{c}{\textbf{DriftQL}} \\
\cmidrule(lr){2-3} \cmidrule(lr){4-5} \cmidrule(lr){6-7}
\textbf{Dataset} & \textbf{Reported} & \textbf{Ours} & \textbf{Reported} & \textbf{Ours} & \textbf{Reported} & \textbf{Ours} \\
\midrule
antmaze-umaze-v2 & 98.5 & 97.3 & 96.8 & 98.3 & 96 & 95.1 \\
antmaze-umaze-diverse-v2 & 93.5 & 92.7 & 91.9 & 87.9 & 86 & 87.6 \\
antmaze-medium-play-v2 & 88.0 & 79.7 & 81.9 & 82.9 & 81 & 75.7 \\
antmaze-medium-diverse-v2 & 85.0 & 69.7 & 61.6 & 62.9 & 75 & 73.0 \\
antmaze-large-play-v2 & 90.0 & 86.2 & 82.6 & 82.9 & 83 & 78.2 \\
antmaze-large-diverse-v2 & 88.0 & 86.5 & 84.1 & 84.9 & 84 & 54.0 \\
\midrule
\textbf{Average} & 90.5 & 85.3 & 83.1 & 83.3 & 84.2 & 77.3 \\
\bottomrule
\end{tabular}
\end{adjustbox}

\end{table}

\begin{table}[!htbp]
\caption{D4RL Adroit baseline reproduction table. Reported values are taken from the corresponding papers; ``Ours'' denotes evaluation under the ReBRAC-v2 final-checkpoint protocol.}
\label{tab:d4rl-adroit-reproductions}
\centering
\appendixtableformat
\setlength{\tabcolsep}{4pt}
\begin{adjustbox}{max width=\linewidth}
\begin{tabular}{@{}lrrrrrr@{}}
\toprule
& \multicolumn{2}{c}{\textbf{FAC}} & \multicolumn{2}{c}{\textbf{GFP}} & \multicolumn{2}{c}{\textbf{DriftQL}} \\
\cmidrule(lr){2-3} \cmidrule(lr){4-5} \cmidrule(lr){6-7}
\textbf{Dataset} & \textbf{Reported} & \textbf{Ours} & \textbf{Reported} & \textbf{Ours} & \textbf{Reported} & \textbf{Ours} \\
\midrule
pen-human-v1 & 73.9 & 60.5 & 64.6 & 77.0 & 51 & 44.7 \\
pen-cloned-v1 & 103.2 & 95.1 & 77.1 & 81.0 & 63 & 53.6 \\
door-human-v1 & 5.5 & 2.6 & 0.3 & 0.1 & 0 & -0.1 \\
door-cloned-v1 & 4.1 & 4.9 & 1.6 & 0.5 & 0 & 0.0 \\
hammer-human-v1 & 8.6 & 4.3 & 4.4 & 1.6 & 1 & 0.4 \\
hammer-cloned-v1 & 11.1 & 7.7 & 12.4 & 9.4 & 1 & 1.0 \\
relocate-human-v1 & 0.6 & 0.1 & 0.5 & 0.3 & 0 & -0.1 \\
relocate-cloned-v1 & 0.5 & 0.2 & 1.6 & 1.5 & 0 & 0.1 \\
\midrule
\textbf{Average} & 25.9 & 21.9 & 20.3 & 21.4 & 14.5 & 12.5 \\
\bottomrule
\end{tabular}
\end{adjustbox}

\end{table}

\paragraph{Matched-Protocol Reproductions.}
\label{sec:matched-baseline-reproductions}
Tables~\ref{tab:ogbench-reproductions}--\ref{tab:d4rl-adroit-reproductions} place reported and matched-protocol scores side by side for FAC, GFP, and DriftQL. All three reproductions preserve their source training horizons while using our final-checkpoint evaluation protocol. For exact reproducibility, we use FAC from \url{https://github.com/JongseongChae/FAC} at commit \texttt{a4f4dd4d5f9d9ecc46392582075da6dd509669b3}, GFP from \url{https://github.com/Simple-Robotics/guided-flow-policy} at commit \texttt{e60468e31066ed072e494be8c6cce14ac19f60e9}, and DriftQL from \url{https://github.com/anashoussaini/driftql} at commit \texttt{1ad1bbc2683b3f7cd8aa6e1d2c04504cceabad94}.

\paragraph{RLiable Construction.}
ReBRAC-v2, FAC, GFP, and DriftQL use per-seed results from our OGBench and D4RL runs. For these methods, RLiable computes stratified-bootstrap uncertainty intervals; aggregate-metric intervals use 50,000 bootstrap replicates. For ReBRAC and FQL, only task-level means are available. We repeat each such task mean across five slots so that RLiable preserves its aggregate point estimate, but this synthetic repetition contains no within-task seed uncertainty; we therefore omit interval estimates for these methods. Consequently, Figure~\ref{fig:rliable-summary} supports aggregate point-estimate comparisons for all included methods, while intervals are informative only for methods with genuine seed-level results. The probability-of-improvement panel uses the same stratified bootstrap and compares ReBRAC-v2 separately against each baseline. 

\FloatBarrier
\section{Ablation Implementation Details}

The grouped regularization ablations in Table~\ref{tab:ablations-full} set the corresponding selected coefficients to zero without changing other hyperparameters. The actor ablation disables behavior-cloning noise, actor-gradient noise, actor weight decay, and normalizing-flow dropout. The critic ablation disables critic dropout, critic-gradient noise, and critic weight decay. The combined ablation disables all seven regularizers. These regularization substitutions do not retune either behavior-cloning coefficient; the separately retuned objective experiments are described in Supplementary Section~\ref{sec:objective-retuning}.

\FloatBarrier
\section{Inference-Time Action Refinement}

\begin{figure}[H]
\centering
\begin{tabular}{@{}ccc@{}}
\includegraphics[width=0.30\textwidth]{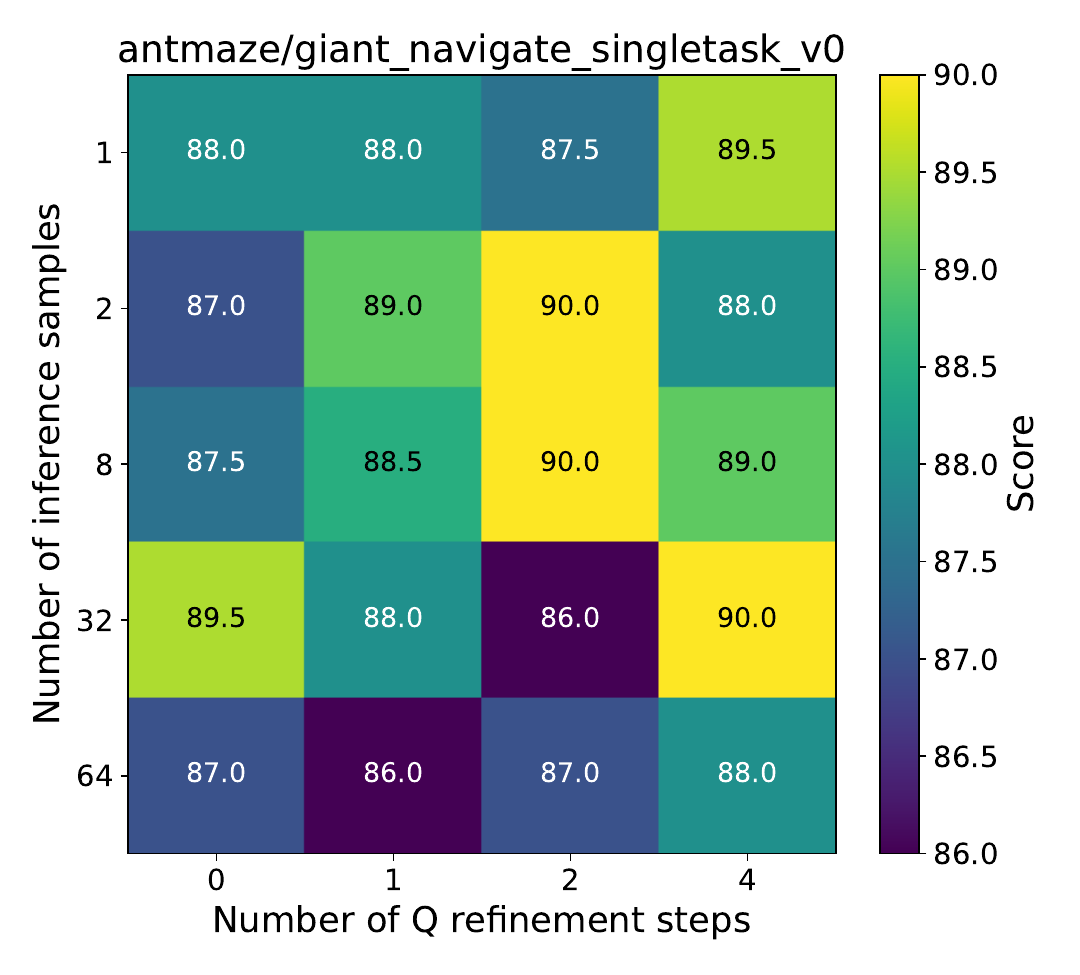} &
\includegraphics[width=0.30\textwidth]{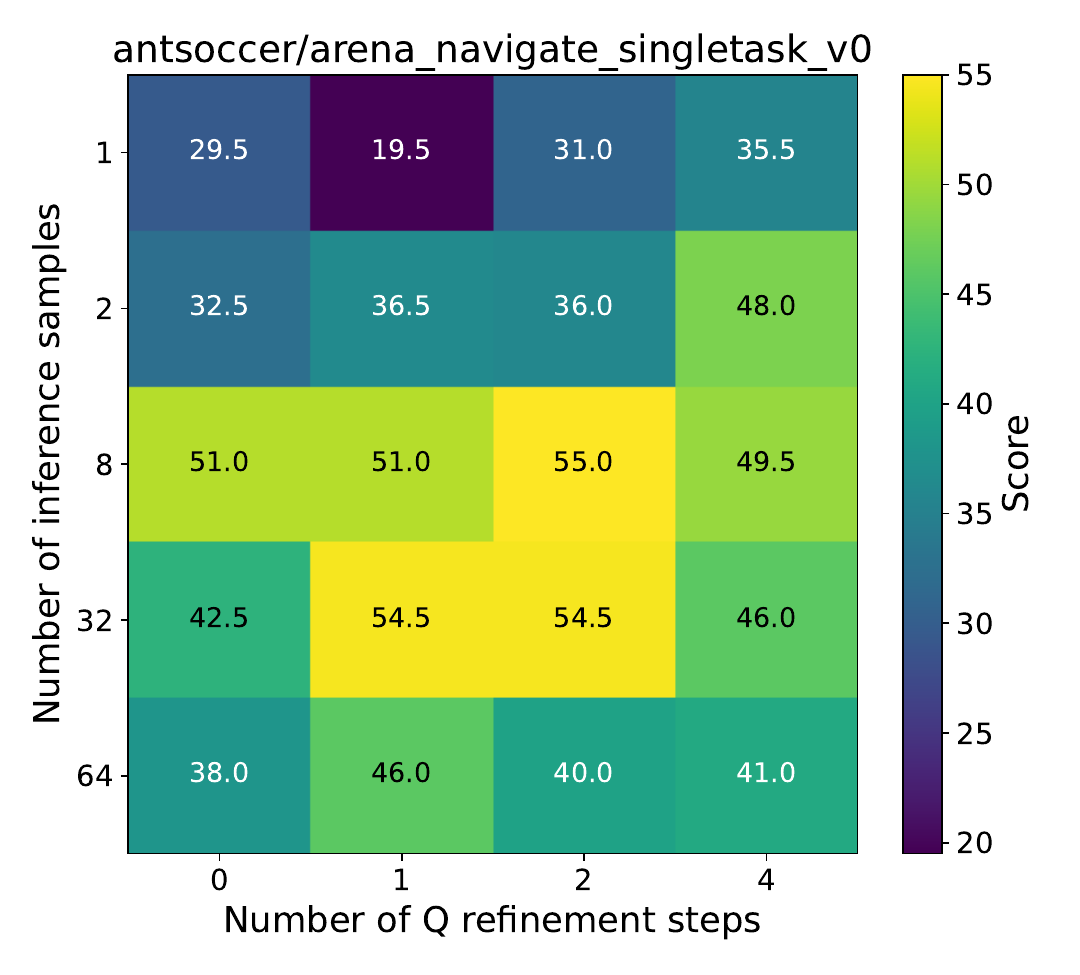} &
\includegraphics[width=0.30\textwidth]{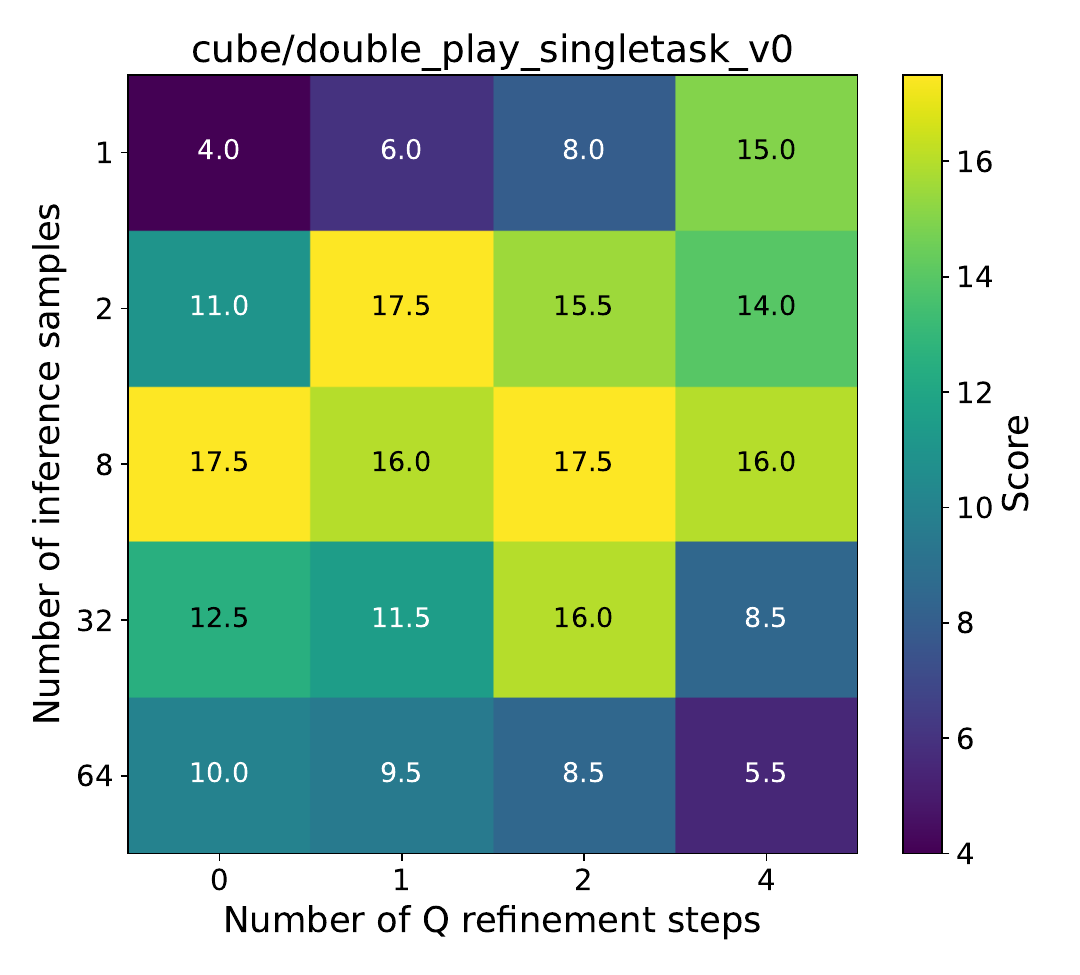}
\\[-2pt]
\small (a) AntMaze Giant Navigate. & \small (b) AntSoccer Arena Navigate. & \small (c) Cube Double Play. \\[4pt]
\includegraphics[width=0.30\textwidth]{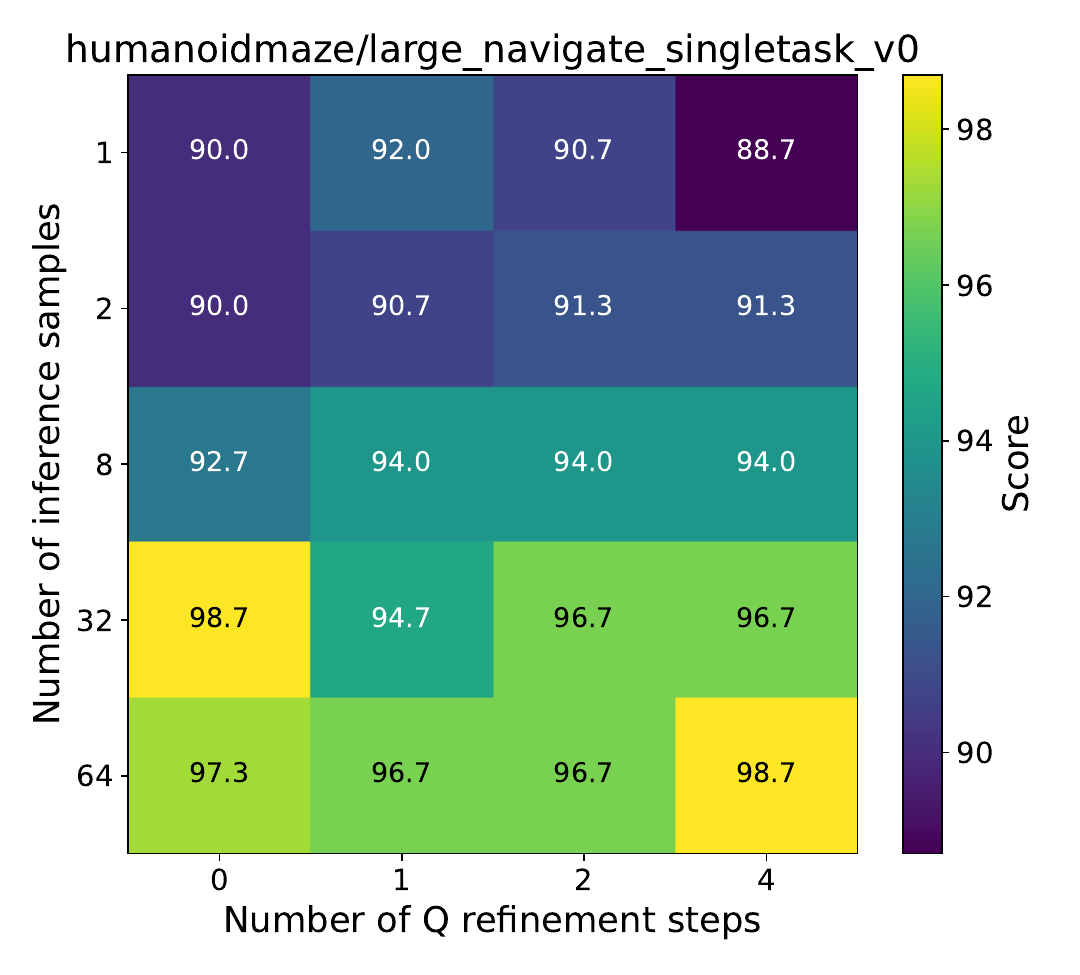} &
\includegraphics[width=0.30\textwidth]{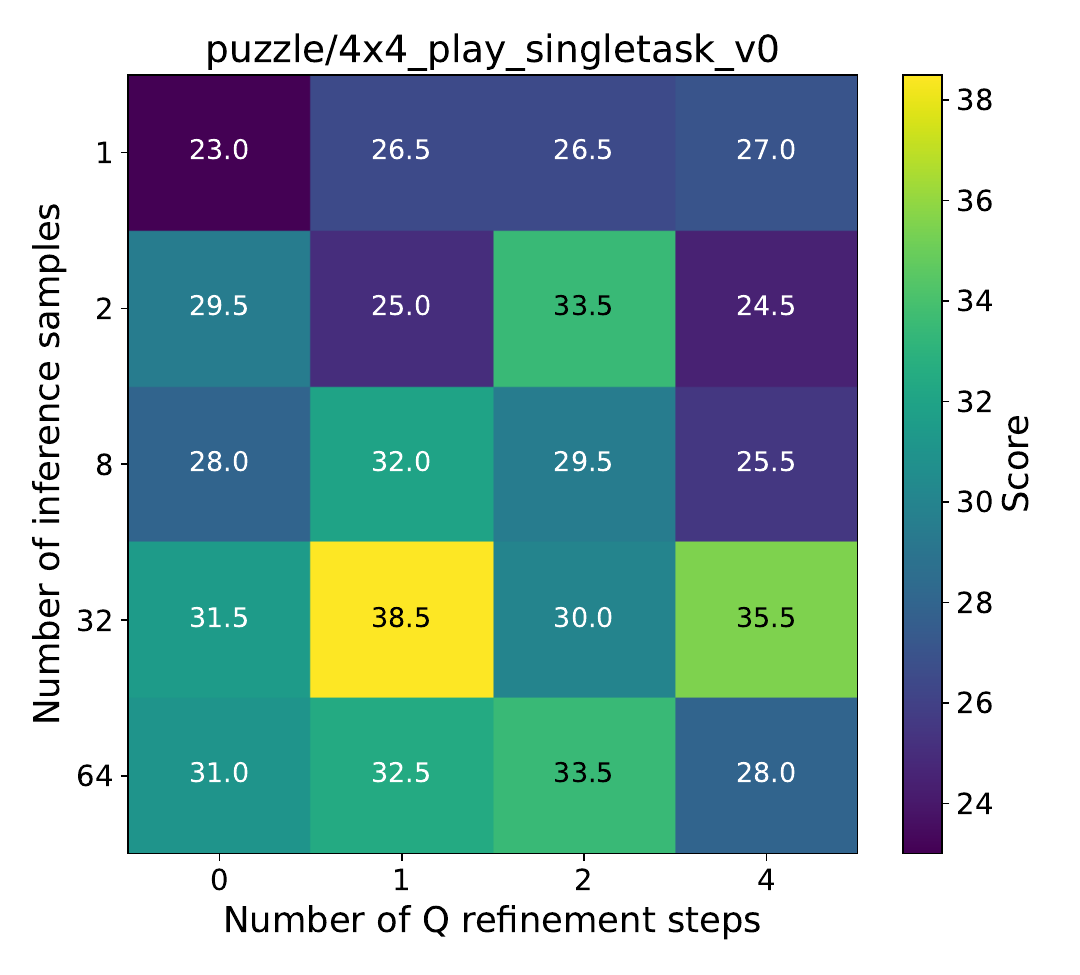} &
\includegraphics[width=0.30\textwidth]{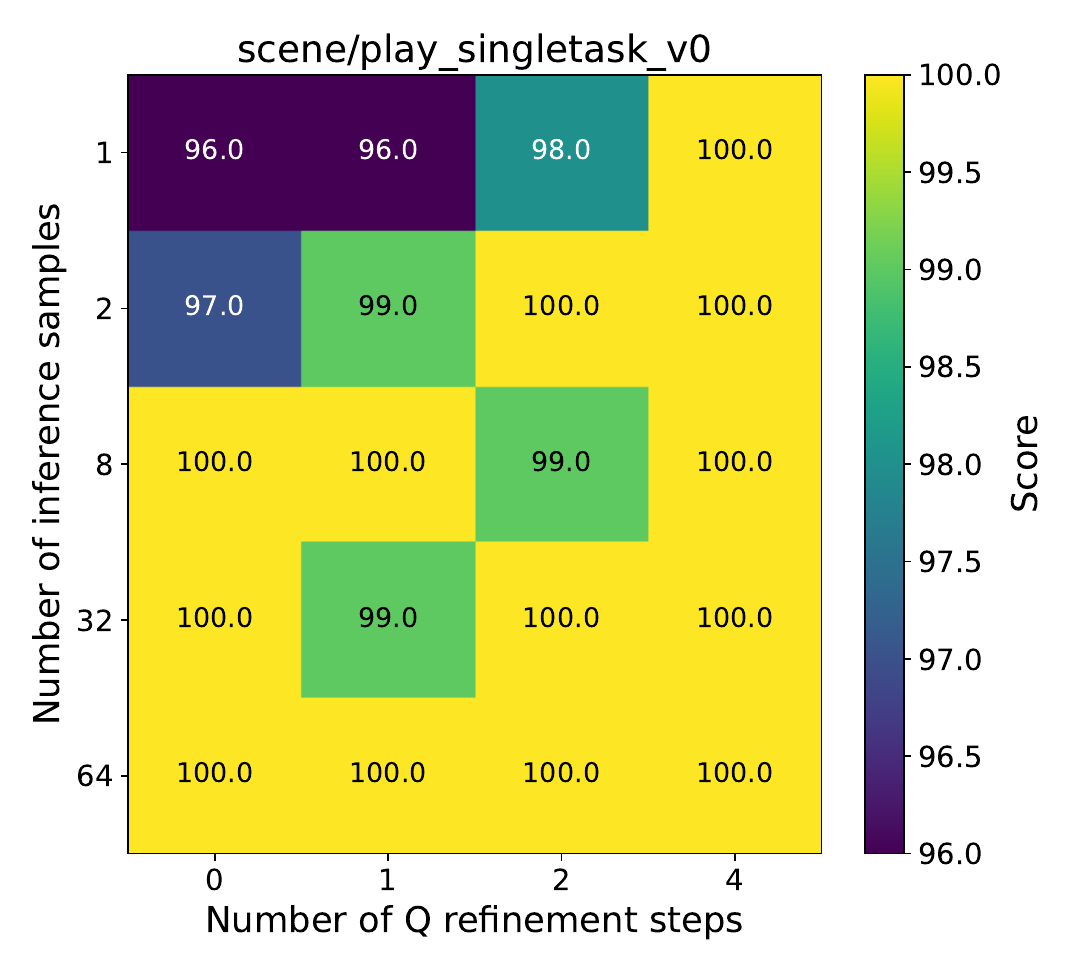} \\
\small (d) HumanoidMaze Large Navigate. & \small (e) Puzzle 4x4 Play. & \small (f) Scene Play.
\end{tabular}
\caption{Task-level sensitivity to inference-time compute. Each heatmap reports normalized score as a function of the number of sampled candidate actions and Q-guided refinement steps.}
\label{fig:inference-grid-tasks}
\end{figure}

Figure~\ref{fig:inference-grid-tasks} provides the six task-level results underlying the aggregate inference-time ablation in Figure~\ref{fig:inference-ablation}. The response to inference compute is heterogeneous. In particular, on AntSoccer Arena, adding one or two refinement steps with 32 candidates increases the score from 42.5 to 54.5, even though refinement produces little change at that sample count after averaging across tasks. The best AntSoccer score, 55.0, occurs with eight candidates and two refinement steps.

\FloatBarrier
\subsection{Additional Inference Compute for Baselines}
\label{sec:baseline-inference-compute}

We apply the same candidate-count and refinement-step grid to FAC, GFP, and DriftQL on the six default tasks used for the ReBRAC-v2 inference ablation: AntMaze Giant, AntSoccer Arena, Cube Double, HumanoidMaze Large, Puzzle $4\times4$, and Scene. Each baseline grid contains four training seeds (10--13) per task, evaluated at the final checkpoint after one million training updates with 50 requested evaluation episodes. We vary the number of policy samples, $K\in\{1,2,8,32,64\}$, and the number of critic-gradient action-refinement steps, $J\in\{0,1,2,4\}$. Each candidate is refined by normalized gradient ascent with step size $0.01$ and clipping to the action bounds, then the candidate with the highest critic score is executed. FAC and GFP use their one-step actors; the existing critic aggregation in each baseline is retained. Training parameters are unchanged across cells, so this experiment tests additional inference computation without retraining the policy for each setting.

\begin{figure}[H]
\centering
\setlength{\tabcolsep}{3pt}
\begin{tabular}{@{}ccc@{}}
\includegraphics[width=0.32\textwidth]{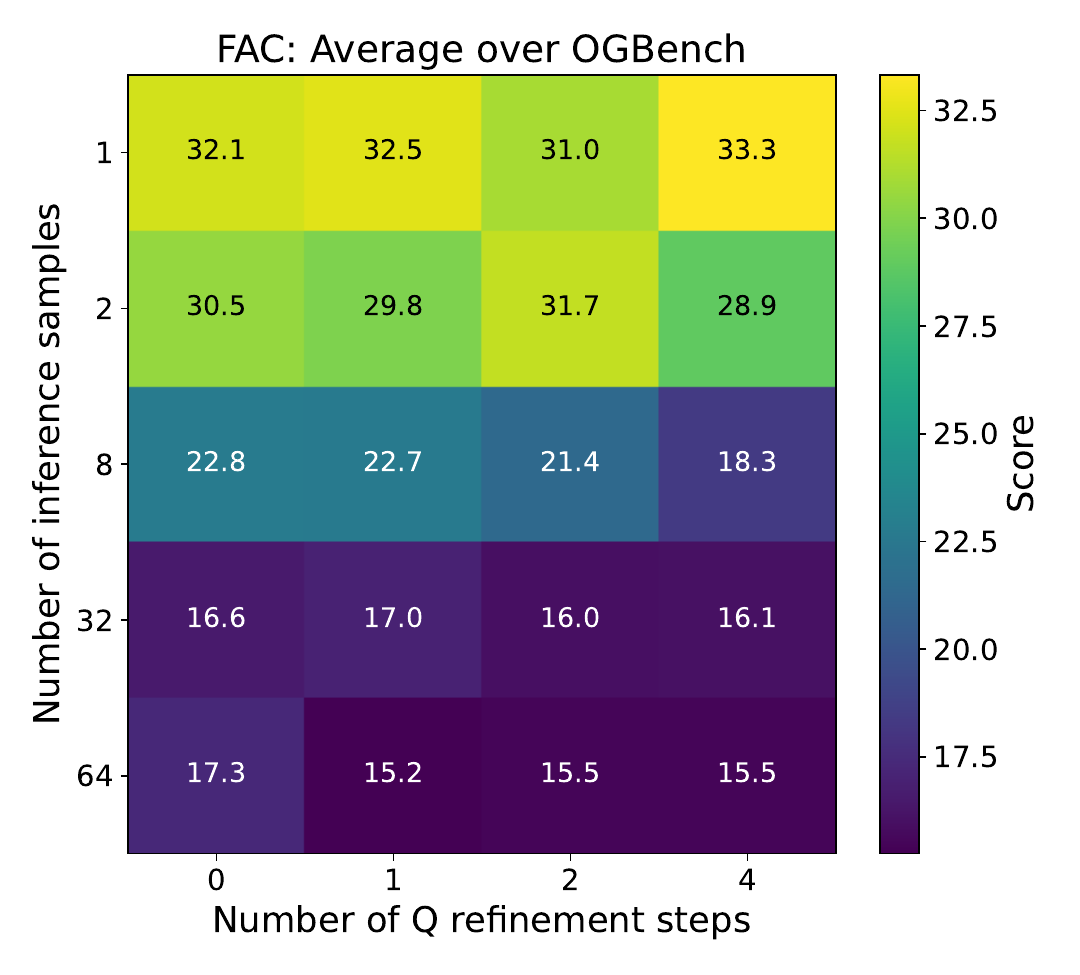} &
\includegraphics[width=0.32\textwidth]{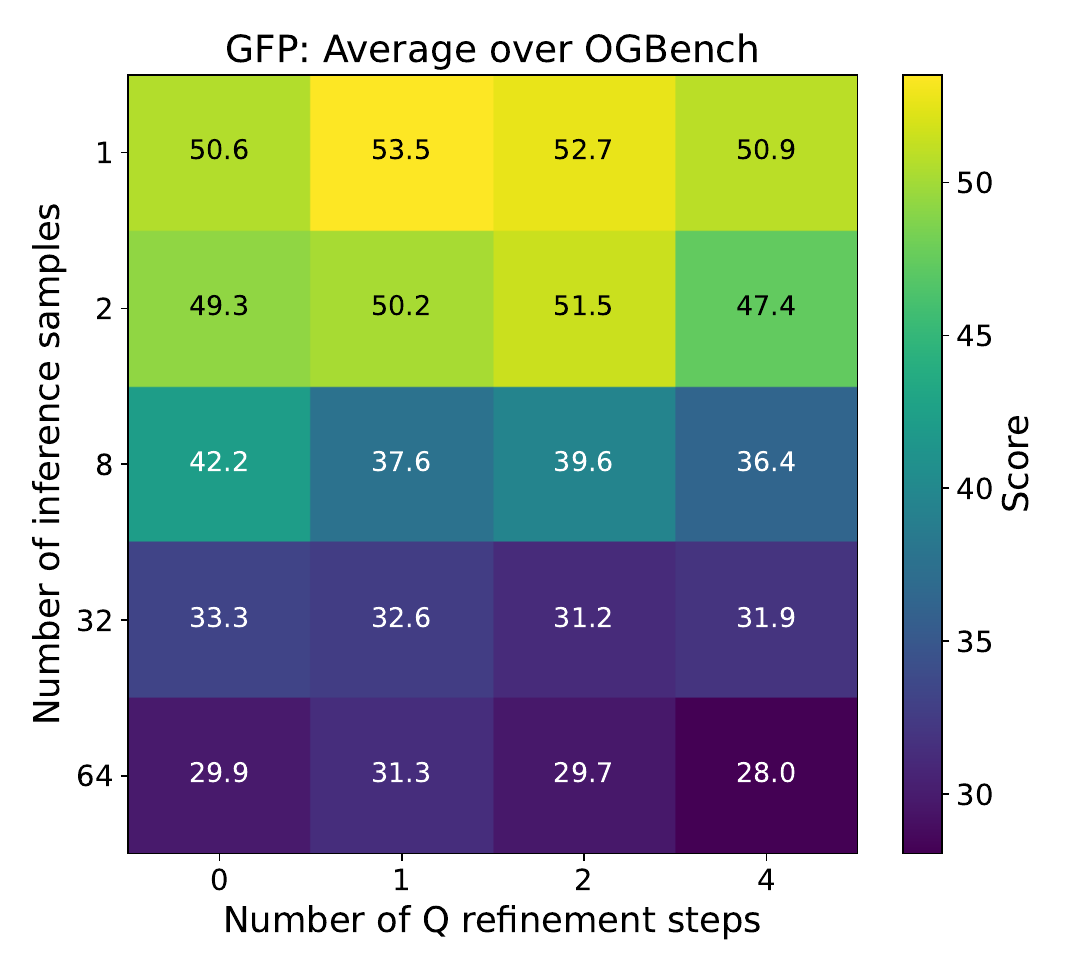} &
\includegraphics[width=0.32\textwidth]{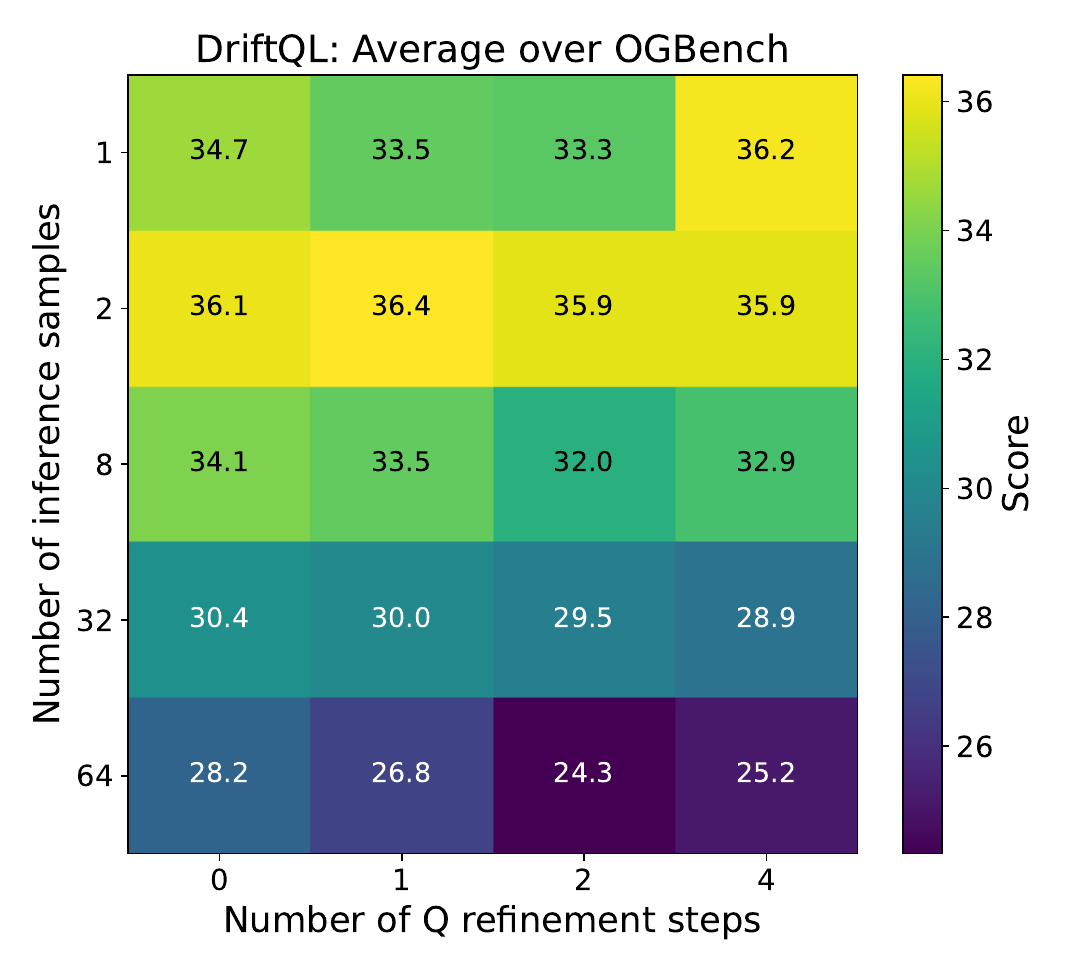}
\end{tabular}
\caption{Additional inference compute for FAC, GFP, and DriftQL. All panels average over the same six default OGBench tasks as Figure~\ref{fig:inference-ablation}, using equal task weights. Color scales differ across panels; numerical scores, rather than colors, should be used for cross-method comparisons.}
\label{fig:baseline-inference-grids}
\end{figure}

\begin{table}[!htbp]
\caption{Inference-compute sensitivity on the same six default OGBench tasks. Each entry is the equally weighted mean of task-level mean normalized scores. Best and worst refer to the 20 tested $(K,J)$ settings; gain is the best score minus the one-action, no-refinement score, computed before rounding. Extrema are descriptive, not significance tests or per-task guarantees.}
\label{tab:inference-compute-comparison}
\centering
\small
\renewcommand{\arraystretch}{1.08}
\setlength{\tabcolsep}{7pt}
\begin{adjustbox}{max width=\linewidth}
\begin{tabular}{@{}lrrrrc@{}}
\toprule
\textbf{Method} & \textbf{One action, no refinement} & \textbf{Worst} & \textbf{Best} & \textbf{Gain} & \textbf{Best $(K,J)$} \\
\midrule
FAC & 32.1 & 15.2 & 33.3 & $+1.3$ & $(1,4)$ \\
GFP & 50.6 & 28.0 & 53.5 & $+3.0$ & $(1,1)$ \\
DriftQL & 34.7 & 24.3 & 36.4 & $+1.8$ & $(2,1)$ \\
ReBRAC-v2 & \textbf{55.1} & \textbf{54.7} & \textbf{64.4} & $+9.3$ & $(32,1)$ \\
\bottomrule
\end{tabular}
\end{adjustbox}

\end{table}

The baselines benefit only modestly from the tested modifications: their best grid settings improve on one action without refinement by 1.3 points for FAC, 3.0 for GFP, and 1.8 for DriftQL. None improves monotonically with more candidates; using 32 or 64 candidates substantially reduces their average scores. In contrast, ReBRAC-v2 gains 9.3 points at its best grid setting. Its lowest tested aggregate, 54.7, exceeds the highest baseline aggregate, GFP's 53.5, by 1.1 points before rounding. Even ReBRAC-v2 with one action and no refinement scores 55.1. Thus, within this grid, giving the baselines additional sampling and refinement does not close the gap.

These comparisons concern aggregate point estimates on six tasks, not dominance on every task or a significance test for the 1.1-point margin. The ReBRAC-v2 grid uses the previously reported two to four seeds per task. The experiment does not match wall-clock inference cost or exhaustively retune baseline refinement hyperparameters; it establishes robustness to these concrete inference-time extensions rather than an upper bound on the baselines' achievable performance.

\FloatBarrier
\subsection{Fused Inference Runtime}
\label{sec:inference-runtime}

We measure the computational cost of ReBRAC-v2 action selection with randomly initialized networks on an NVIDIA GeForce RTX 3090. The benchmark uses JAX/jaxlib 0.6.2, Flax 0.10.7, Python 3.10.19, and float32 arithmetic. Each of the ten common OGBench categories uses its released network configuration and actual observation/action dimensions, including the 14-layer flow and four-critic ensemble. We construct the environments only to determine these dimensions; no training, dataset loading, or environment stepping is required.

The measured function jointly compiles flow sampling, optional critic-gradient refinement, and candidate selection. Observations and parameters are already on the GPU, and actions remain there. Each timed call includes dispatch and waits for completion, preventing asynchronous execution from understating latency. This is a fused, device-resident inference implementation, not the complete rollout evaluator; CPU--GPU transfers, observation normalization, logging, and environment execution are excluded. Thus these measurements characterize inference computation, not end-to-end control-loop latency or baseline-relative runtime.

We measure one state per call, matching sequential action selection in a single environment. We compare one action without critic evaluation, $(K,J)=(1,0)$; selecting the highest-valued of 32 actions without refinement, $(32,0)$; and the deployed algorithmic setting, $(32,2)$. These 32 candidates correspond to the same state, not 32 different environments. Refinement uses the minimum of the four critic values, normalized action gradients with step size $0.01$, and clipping to the action bounds. Each condition has a separate first call and 20 warmup calls, followed by five repetitions of 100 individually synchronized calls. Compilation and warmup are excluded. Conditions are measured in randomized order within each category. Inputs are drawn from a fixed pool of eight standard-normal state vectors; categorical value bounds are synthetic rather than dataset-derived. Random parameters exercise the inference computation without evaluating policy quality, and timing variability must not be interpreted as uncertainty across trained seeds.

\begin{table}[!htbp]
\caption{Fused, device-resident ReBRAC-v2 inference on an NVIDIA GeForce RTX 3090 for one state per call ($B=1$). Columns $(K,J)$ specify the number of candidates per state and refinement steps per candidate. Input and action dimensions are $d_s$ and $d_a$. Each ratio divides the setting's median latency by the $(1,0)$ median within the same category and batch size. Each condition uses 500 synchronized timed calls after warmup. p95 is the 95th-percentile latency for full inference, $(32,2)$, not a confidence interval. The final row equally averages category-level statistics, including ratios; it is not a pooled percentile. Parameters are randomly initialized; compilation, CPU--GPU transfers, and environment execution are excluded.}
\label{tab:fused-inference-1}
\centering
\small
\setlength{\tabcolsep}{5pt}
\renewcommand{\arraystretch}{1.12}
\begin{adjustbox}{max width=\linewidth}
\begin{tabular}{@{}lrrrrrrrr@{}}
\toprule
 & & & \multicolumn{3}{c}{Median latency (ms)} & Full & \multicolumn{2}{c}{Ratio vs. $(1,0)$} \\
\cmidrule(lr){4-6}\cmidrule(lr){8-9}
Category & $d_s$ & $d_a$ & $(1,0)$ & $(32,0)$ & $(32,2)$ & p95 (ms) & $(32,0)$ & $(32,2)$ \\
\midrule
AntMaze Large & 29 & 8 & 1.173 & 1.211 & 1.583 & 1.616 & $1.03\times$ & $1.35\times$ \\
AntMaze Giant & 29 & 8 & 1.160 & 1.195 & 1.587 & 1.627 & $1.03\times$ & $1.37\times$ \\
HumanoidMaze Medium & 69 & 21 & 1.138 & 1.237 & 1.584 & 1.612 & $1.09\times$ & $1.39\times$ \\
HumanoidMaze Large & 69 & 21 & 1.147 & 1.250 & 1.575 & 1.642 & $1.09\times$ & $1.37\times$ \\
AntSoccer Arena & 42 & 8 & 1.173 & 1.201 & 1.574 & 1.605 & $1.02\times$ & $1.34\times$ \\
Cube Single & 28 & 5 & 1.064 & 1.201 & 1.576 & 1.596 & $1.13\times$ & $1.48\times$ \\
Cube Double & 37 & 5 & 1.071 & 1.191 & 1.603 & 1.684 & $1.11\times$ & $1.50\times$ \\
Scene & 40 & 5 & 1.067 & 1.196 & 1.589 & 1.636 & $1.12\times$ & $1.49\times$ \\
Puzzle $3\times3$ & 55 & 5 & 1.078 & 1.184 & 1.600 & 1.624 & $1.10\times$ & $1.48\times$ \\
Puzzle $4\times4$ & 83 & 5 & 1.072 & 1.208 & 1.598 & 1.621 & $1.13\times$ & $1.49\times$ \\
\midrule
\textbf{Mean over categories} & --- & --- & 1.114 & 1.207 & 1.587 & 1.626 & $1.08\times$ & $1.43\times$ \\
\bottomrule
\end{tabular}
\end{adjustbox}

\end{table}

For single-state inference, the mean of the ten category-level median latencies is 1.114\,ms for actor-only sampling, 1.207\,ms for best-of-32 selection, and 1.587\,ms for sampling plus refinement. Relative to actor-only inference, best-of-32 selection adds 0.093\,ms on average, while sampling plus refinement adds 0.472\,ms. Their mean within-category latency ratios are $1.08\times$ and $1.43\times$, respectively (8.5\% and 42.6\% overhead, computed before rounding). Each ratio uses the same category's single-action, no-refinement median as its denominator. The p95 column reports the 95th percentile of timed full-inference calls, describing upper-tail latency rather than statistical confidence. Full-inference medians range from 1.574 to 1.603\,ms across categories. The small variation across these dimensions is specific to this architecture, device, and execution regime; it does not imply dimension-independent computational complexity.

These results quantify the absolute inference cost of the selected recipe without implying that its overhead is negligible or that other methods are slower.

\FloatBarrier
\clearpage
\section{Normalizing-Flow Depth Ablation}

Table~\ref{tab:nf-depth} provides the task-level results corresponding to Figure~\ref{fig:nf-depth-ablation}.

\begin{table}[H]
\caption{Effect of the normalizing-flow depth (number of coupling layers). Cells are shaded green/red when the score is above/below the reference depth.}
\label{tab:nf-depth}
\centering
\footnotesize
\setlength{\tabcolsep}{2pt}

\resizebox{\linewidth}{!}{%
\begin{tabular}{lccccccc}
\toprule
Ablation & \shortstack{AntMaze\\Giant} & \shortstack{AntSoccer\\Arena} & \shortstack{Cube\\Double} & \shortstack{HumanoidMaze\\Large} & \shortstack{Puzzle\\4x4} & scene & Avg. \\
\midrule
depth 2 & \cellcolor{red!12}3.0 $\pm$ 3.8 & \cellcolor{red!12}18.5 $\pm$ 9.4 & \cellcolor{red!12}0.5 $\pm$ 1.0 & \cellcolor{red!12}55.5 $\pm$ 6.4 & \cellcolor{red!12}17.5 $\pm$ 10.2 & \cellcolor{red!12}92.0 $\pm$ 8.5 & 31.2 \\
depth 4 & \cellcolor{red!12}26.5 $\pm$ 13.9 & \cellcolor{red!12}35.0 $\pm$ 10.1 & \cellcolor{red!12}6.0 $\pm$ 5.2 & \cellcolor{red!12}64.5 $\pm$ 16.8 & \cellcolor{red!12}28.5 $\pm$ 3.4 & 100.0 $\pm$ 0.0 & 43.4 \\
depth 8 & \cellcolor{red!12}82.0 $\pm$ 7.7 & \cellcolor{green!12}51.5 $\pm$ 11.4 & \cellcolor{red!12}7.5 $\pm$ 6.0 & \cellcolor{red!12}91.0 $\pm$ 5.8 & \cellcolor{green!12}37.0 $\pm$ 10.4 & \cellcolor{red!12}99.0 $\pm$ 1.2 & 61.3 \\
depth 12 & \cellcolor{green!12}91.0 $\pm$ 3.5 & \cellcolor{red!12}48.0 $\pm$ 10.7 & \cellcolor{red!12}11.5 $\pm$ 3.8 & \cellcolor{red!12}95.0 $\pm$ 4.8 & \cellcolor{red!12}21.0 $\pm$ 4.2 & 100.0 $\pm$ 0.0 & 61.1 \\
depth 14 (ref.) & 88.0 $\pm$ 1.6 & 49.0 $\pm$ 10.0 & 14.5 $\pm$ 8.4 & 97.0 $\pm$ 1.2 & 33.5 $\pm$ 15.0 & 100.0 $\pm$ 0.0 & 63.7 \\
depth 18 & \cellcolor{green!12}95.5 $\pm$ 2.5 & \cellcolor{green!12}60.0 $\pm$ 3.3 & \cellcolor{red!12}12.0 $\pm$ 4.9 & 97.0 $\pm$ 2.6 & \cellcolor{red!12}27.5 $\pm$ 9.0 & 100.0 $\pm$ 0.0 & 65.3 \\
\bottomrule
\end{tabular}}
\end{table}

\end{document}